\documentclass[10pt,twocolumn,letterpaper]{article}
\pdfoutput=1
\usepackage[pagenumbers]{cvpr} 
\usepackage[dvipdfmx]{graphicx} 
\usepackage{bmpsize}
\usepackage{amsmath}
\usepackage{amssymb}
\usepackage{booktabs}
\usepackage{multirow}
\usepackage{bm}
\usepackage{soul}
\usepackage{algorithm}
\usepackage{algpseudocode}
\usepackage{enumitem}
\usepackage{overpic}
\usepackage[accsupp]{axessibility} 

\usepackage[pagebackref,breaklinks,colorlinks]{hyperref}

\newcommand{\ourmethod}{UDPReg\xspace}

\usepackage[pagebackref,breaklinks,colorlinks]{hyperref}

\usepackage[capitalize]{cleveref}
\crefname{section}{Sec.}{Secs.}
\Crefname{section}{Section}{Sections}
\Crefname{table}{Table}{Tables}
\crefname{table}{Tab.}{Tabs.}

\usepackage{caption}
\usepackage{enumitem}
\setitemize{noitemsep,topsep=0pt,parsep=0pt,partopsep=0pt}

\def\cvprPaperID{1458} 
\def\confName{CVPR}
\def\confYear{2023}

\begin{document}

\title{Unsupervised Deep Probabilistic Approach for Partial Point Cloud Registration}

\author{
Guofeng Mei$^1$ \quad Hao Tang$^2$ \quad Xiaoshui Huang$^3$ \quad Weijie Wang$^4$ \\ 
Juan Liu$^5$ \quad Jian Zhang$^1$ \quad Luc Van Gool$^2$ \quad Qiang Wu$^1$\\
$^1$GBDTC,UTS \quad $^2$CVL, ETH Zurich  \quad $^3$Shanghai AI Lab \quad $^4$UNITN \quad $^5$BIT \\
{\tt\small Guofeng.Mei@student.uts.edu.au}
\quad {\tt\small liu{\_}juan@bit.edu.au}
}
\maketitle

\begin{abstract}
Deep point cloud registration methods face challenges to partial overlaps and rely on labeled data.
To address these issues, we propose \ourmethod, an unsupervised deep probabilistic registration framework for point clouds with partial overlaps. Specifically, we first adopt a network to learn posterior probability distributions of Gaussian mixture models (GMMs) from point clouds. 
To handle partial point cloud registration, we apply the Sinkhorn algorithm to predict the distribution-level correspondences under the constraint of the mixing weights of GMMs.
To enable unsupervised learning, we design three distribution consistency-based losses: self-consistency, cross-consistency, and local contrastive. 
The self-consistency loss is formulated by encouraging GMMs in Euclidean and feature spaces to share identical posterior distributions. 
The cross-consistency loss derives from the fact that the points of two partially overlapping point clouds  belonging to the same clusters share the cluster centroids. 
The cross-consistency loss allows the network to flexibly learn a transformation-invariant posterior distribution of two aligned point clouds. 
The local contrastive loss facilitates the network to extract discriminative local features.
Our \ourmethod achieves competitive performance on the 3DMatch/3DLoMatch and ModelNet/ModelLoNet benchmarks.
\end{abstract}

\vspace{-0.3cm}
\section{Introduction}\label{sec:intro}
Rigid point cloud registration aims at determining the optimal transformation to align two partially overlapping point clouds into one coherent coordinate system~\cite{huang2020feature,mei2022overlap,mei2021point,mei2022partial}. This task dominates the performance of systems in many areas, such as robotics~\cite{Zhou2022}, augmented reality~\cite{borrmann2018large}, autonomous driving~\cite{nagy2018real,wang2019robust}, radiotherapy~\cite{li2019noninvasive}, etc. 
Recent advances have been monopolized by learning-based approaches due to the development of 3D point cloud representation learning and differentiable optimization~\cite{qin2022geometric}. 

Existing deep learning-based point cloud registration methods can be broadly categorized as \textit{correspondence-free}~\cite{aoki2019pointnetlk,huang2020feature,mei2021point,mei2022partial,xu2021omnet} and \textit{correspondence-based}~\cite{choy2020deep,bai2021pointdsc,huang2021predator,yew2022regtr}.
The former minimizes the difference between global features extracted from two input point clouds.
These global features are typically computed based on all the points of a point cloud, making correspondence-free approaches inadequate to handle real scenes with partial overlap~\cite{zhang2020deep,choy2020deep}. 
Correspondence-based methods first extract local features used for the establishment of point-level~\cite{choy2020deep,huang2020feature,huang2021predator,fu2021robust} or distribution-level~\cite{magnusson2009evaluation,stoyanov2012point,evangelidis2017joint,yuan2020deepgmr} correspondences, and finally, estimate the pose from those correspondences.
However, point-level registration does not work well under conditions involving varying point densities or repetitive patterns~\cite{mei2022overlap}. 
This issue is especially prominent in indoor environments, where low-texture regions or repetitive patterns sometimes dominate the field of view.
Distribution-level registration, which compensates for the shortcomings of point-level methods, aligns two point clouds without establishing explicit point correspondences. 
Unfortunately, to the best of our knowledge, the existing methods are inflexible and cannot handle point clouds with partial overlaps in real scenes~\cite{mei2022overlap,li2022gaussian}.  
Moreover, the success of learning-based methods mainly depends on large amounts of ground truth transformations or correspondences as the supervision signal for model training. 
Needless to say, the required ground truth is typically difficult or costly to acquire, thus hampering their application in the real world~\cite{shen2022reliable}. 

We thus propose an unsupervised deep probabilistic registration framework to alleviate these limitations. 
Specifically, we extend the distribution-to-distribution (D2D) method to solve partial point cloud registration by adopting the Sinkhorn algorithm~\cite{cuturi2013sinkhorn} to predict correspondences of distribution.
In order to make the network learn geometrically and semantically consistent features, we design distribution-consistency losses, i.e., self-consistency and cross-consistency losses, to train the networks without using any ground-truth pose or correspondences. Besides, we also introduce a local contrastive loss to learn more discriminative features by pushing features of points belonging to the same clusters together while pulling dissimilar features of points coming from different clusters apart. 

Our \ourmethod is motivated by OGMM~\cite{mei2023overlap} and UGMM~\cite{huang2022unsupervised} but differs from them in several ways. Firstly, unlike OGMM, which is a supervised method, our approach is unsupervised. Secondly, while UGMM~\cite{huang2022unsupervised} treats all clusters equally in the matching process, our method aligns different clusters with varying levels of importance. This enables our approach to handle partial point cloud registration successfully.
To enable unsupervised learning, the designed self-consistency loss encourages the extracted features to be geometrically consistent by compelling the features and coordinates to share the posterior probability. The cross-consistency loss prompts the extracted features to be geometrically consistent by forcing the partially overlapping point clouds to share the same clusters.  
We evaluate our \ourmethod on 3DMatch\cite{zeng20173dmatch}, 3DLoMatch\cite{huang2021predator}, ModelNet\cite{wu20153d} and ModelLoNet\cite{huang2021predator}, comparing our approach against traditional and deep learning-based point cloud registration approaches.
\ourmethod achieves state-of-the-art results and significantly outperforms unsupervised methods on all the benchmarks.

In summary, the main contributions of this work are:
\begin{itemize} [leftmargin=*]
    \item We propose an unsupervised learning-based probabilistic framework to register point clouds with partial overlaps.
    \item We provide a deep probabilistic framework to solve partial point cloud registration by adopting the Sinkhorn algorithm to predict distribution-level correspondences.
    \item We formulate self-consistency, cross-consistency, and local-contrastive losses, to make the posterior probability in coordinate and feature spaces consistent so that the feature extractor can be trained in an unsupervised way.
    \item We achieve state-of-the-art performance on a comprehensive set of experiments, including synthetic and real-world datasets\footnote{\url{https://github.com/gfmei/UDPReg}}.
\end{itemize}

\begin{figure*}[!hbt]
\centering
\includegraphics[width=1.0\textwidth]{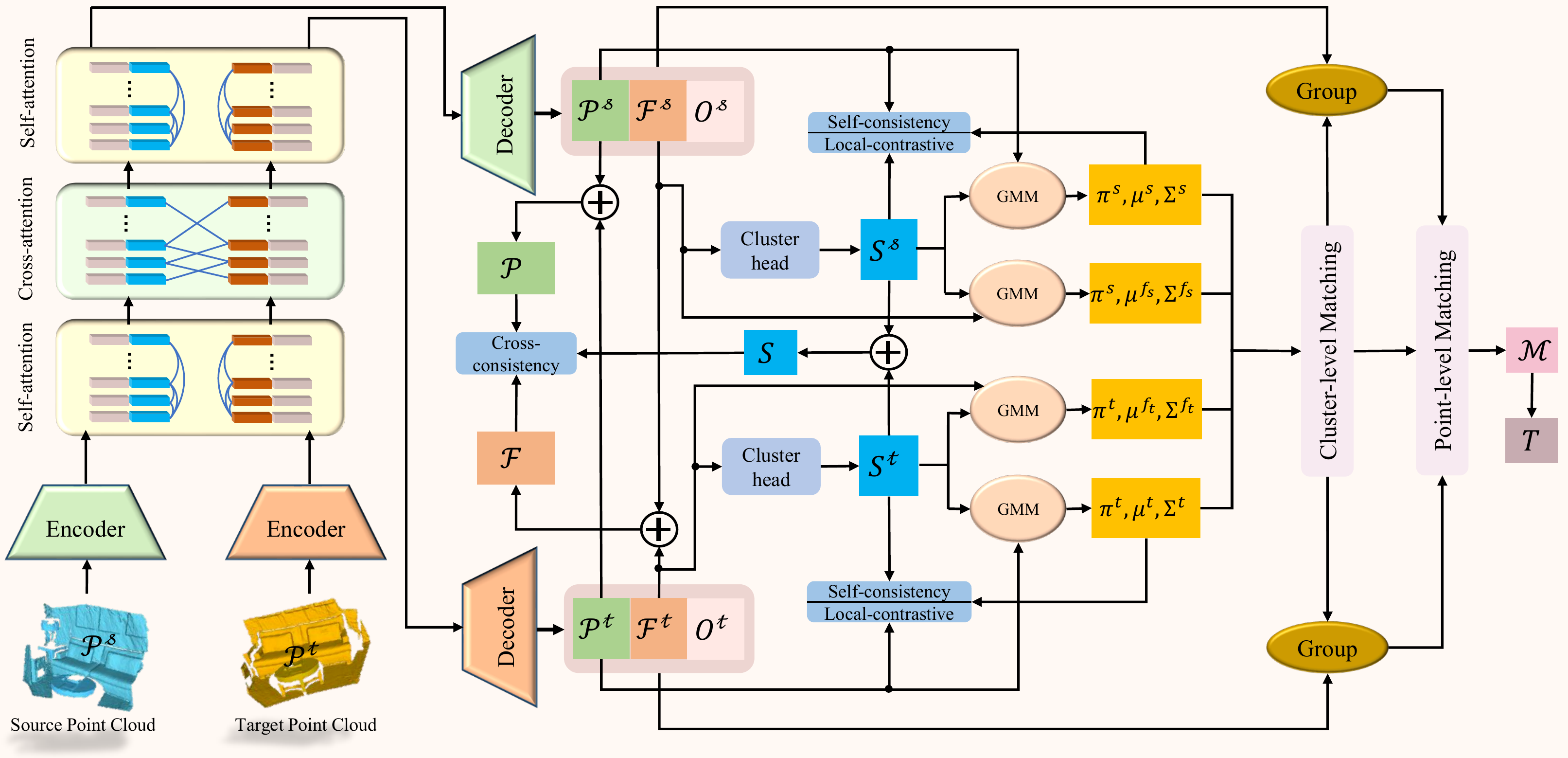}
\caption{
\ourmethod uses shared weighted network extracts point-level features $\bm{\mathcal{F}}^s$ and $\bm{\mathcal{F}}^t$, overlap scores $\bm{O}^s$ and $\bm{O}^t$ from point clouds $\bm{\mathcal{P}}^s$ and $\bm{\mathcal{P}}^t$, respectively.
Cluster head consumes $\bm{\mathcal{F}}^s$ and $\bm{\mathcal{F}}^t$ to calculate probability matrices $\bm{S}^s$ and $\bm{S}^t$, which are applied to estimate the parameters $(\pi^x_{j},\bm{\mu}^x_{j},\bm{\Sigma}^x_{j}), x{\in}\{s,t,f_s,f_t\}$, of GMMs. Next, cluster-level and point-level matching modules estimate the correspondences $\mathcal{M}$, which are used to estimate the transformation $T$. The network is trained using local contrastive, self-consistency, and cross-consistency losses. $\bm{S}$ is the concatenation of $\bm{S}^s$ and $\bm{S}^t$.
$\bm{\mathcal{P}}$ and $\bm{\mathcal{F}}$ are the concatenation of $\bm{\mathcal{P}}^s$ and $\bm{\mathcal{P}}^t$, and $\bm{\mathcal{F}}^s$ and $\bm{\mathcal{F}}^t$, respectively.}
\vspace{-0.4cm}
\label{fig:gmm}
\end{figure*}

\section{Related Work}

\noindent \textbf{Point-Level Methods.}
Point-level registration approaches first extract point-wise features, then establish point-to-point correspondences through feature matching, followed by outlier rejection and robust estimation of the rigid transformation.
Numerous works, such as FCGF~\cite{choy2019fully} and RGM~\cite{fu2021robust}, focus on extracting discriminative features for geometric correspondences. 
For the correspondence prediction,
DCP~\cite{wang2019deep}, RPMNet \cite{yew2020rpm}, and REGTR~\cite{yew2022regtr}  perform feature matching by integrating the Sinkhorn algorithm or Transformer~\cite{vaswani2017attention} into a network to generate soft correspondences from local features. IDAM \cite{li2019iterative} incorporates both geometric and distance features into the iterative matching process. 
To reject outliers, DGR~\cite{choy2020deep} and 3DRegNet~\cite{pais20203dregnet} use networks to estimate the inliers.  Predator~\cite{huang2021predator} and PRNet~\cite{wang2019prnet} focus on detecting points in the overlap region and utilizing their features to generate matches.  Keypoint-free methods~\cite{mei2021point,zhang2022patchformer,yu2021cofinet} first downsample the point clouds into super-points and then match them by examining whether their neighborhoods (patch) overlap. Though achieving remarkable performance, most of these methods rely on large amounts of ground-truth transformations, as inaccessible or expensive as such annotation may get. 
This said, the ground-truth geometric labels could potentially be obtained from full 3D reconstruction pipelines~\cite{choi2015robust},
but these require delicate parameter tuning, partial human supervision, and extra sensory information such as GPS.
As a result, the success of learning-based techniques has been limited to a handful of datasets with ground-truth annotations. 

\noindent \textbf{Distribution-Level Methods.}
Distribution-level methods model the point clouds as probability distributions, often via the use of GMMs, and perform alignment either by employing a correlation-based or an EM-based optimization framework.
The correlation-based methods~\cite{jian2010robust,yuan2020deepgmr} first build GMM probability distributions for both the source and target point clouds. Then, the transformation is estimated by minimizing a metric or divergence between the distributions.  However, these methods lead to nonlinear optimization problems with nonconvex constraints~\cite{lawin2018density}. Unlike correlation-based methods, the
EM-based approaches, such as JRMPC~\cite{evangelidis2017joint}, CPD~\cite{myronenko2010point}, and FilterReg \cite{gao2019filterreg}, represent the geometry of one point cloud using a GMM distribution over 3D Euclidean space. 
The transformation is then calculated by fitting another point cloud to the GMM distribution under the maximum likelihood estimation (MLE) framework.
These methods are robust to noise and density variation~\cite{yuan2020deepgmr}. Most of them utilize robust discrepancies to reduce the influence of outliers by greedily aligning the largest possible fraction of points while being tolerant to a small number of outliers. However, if outliers dominate, the greedy behavior of these methods easily emphasizes outliers, leading to degraded registration results~\cite{evangelidis2017joint}.
Considering these factors, we formulate registration in a novel partial distribution matching framework, where we only seek to partially match the distributions.

\noindent \textbf{Unsupervised Point Cloud Registration.}
To handle ground-truth labeling issues, great efforts~\cite{deng2018ppf,huang2020feature,yang2021self,jiang2021sampling,shen2022reliable,wang2019prnet} have been devoted to unsupervised deep point cloud registration. The existing methods mainly lie in auto-encoders~\cite{deng2018ppf,huang2020feature,shen2022reliable} with a reconstruction loss or contrastive learning \cite{xie2020pointcontrast,el2021unsupervisedr,choy2019fully} with data augmentation.
Although encouraging results have been achieved, some limitations remain to be addressed. 
Firstly, they depend on the point-level loss, such as Chamfer distance in auto-encoder~\cite{deng2018ppf}, finding it difficult to handle large-scale scenarios due to computational complexity.
Secondly, many pipelines~\cite{wang2019prnet} apply fixed/handcrafted data augmentation to generate transformations or correspondences, leading to sub-optimal learning. 
This is because they cannot fully use the cross information of partially overlapping point clouds without geometric labels, and the shape complexity of the samples is ignored in the fixed augmentation~\cite{li2022hybridcr}.
To overcome these limitations, we provide a distribution consistency-based unsupervised method, which utilizes the distribution-level loss to reduce the computational complexity. Even without using any data augmentation, the proposed method is still suitable and available. 


\section{Method}
\subsection{Problem Formulation}
Point cloud registration aims to seek a transformation $T{\in} SE(3)$ that optimally aligns the source point cloud $\bm{\mathcal{P}}^s{=}\{\bm{p}^s_i {\in}\mathbb{R}^{3}\big|i {=} 1, 2, ..., N_s\}$ to the target point cloud $\bm{\mathcal{P}}^t{=}\{\bm{p}^t_j {\in}\mathbb{R}^{3}\big|j {=} 1, 2, ..., N_t\}$.  $\bm{\mathcal{P}}^s$ and $\bm{\mathcal{P}}^t$ contain $N_s$ and $N_t$ points, respectively. $T$ consists of rotation $R{\in} SO(3)$ and translation $\bm{t}{\in} \mathbb{R}^3$.  
Instead of directly employing the point-level solution, we apply the distribution-to-distribution (D2D) approach to fit these two point clouds and obtain individual potential GMMs, where each component represents the density of the spatial coordinates and features in a local region. The transformation is then recovered from the learned GMMs. Our goal is to learn GMMs of point clouds for registration without any ground-truth geometric labels. 
Our \ourmethod framework is conceptually simple and is illustrated in Fig.~\ref{fig:gmm}.
The shared weighted feature extractor consisting of an encoder, Transformer (self- and cross-attention), and decoder first extracts point-wise features $\bm{\mathcal{F}}^s$ and $\bm{\mathcal{F}}^t$, overlap scores $\bm{O}^s$ and $\bm{O}^t$ from point clouds $\bm{\mathcal{P}}^s$ and $\bm{\mathcal{P}}^t$, respectively. 
$\bm{\mathcal{F}}^s$ and $\bm{\mathcal{F}}^t$ are then fed to cluster head to estimate the distributions (GMMs) of $\bm{\mathcal{P}}^s$ and $\bm{\mathcal{P}}^t$ in both coordinate and feature spaces. 
After that, the correspondences $\mathcal{M}$ are estimated by performing cluster-level and point-level matching based on the Sinkhorn algorithm \cite{cuturi2013sinkhorn}.
Finally, a variant of RANSAC \cite{fischler1981random} specialized to 3D registration is adopted to calculate $T$ based on the estimated correspondences.
The network is trained using the proposed self-consistency, cross-consistency, and local contrastive losses in an unsupervised manner.


\subsection{The Proposed GMM-Based Registration}\label{subs:point}
\noindent\textbf{Feature Extraction.}
Following~\cite{huang2021predator,qin2022geometric,mei2022overlap}, a shared encoder KPConv-FPN~\cite{thomas2019kpconv}, which is composed of a series of ResNet-like blocks and stridden convolutions, simultaneously downsamples the raw point clouds $\bm{\mathcal{P}}^s$ and $\bm{\mathcal{P}}^t$ into superpoints $\bar{\bm{\mathcal{P}}}^s$ and $\bar{\bm{\mathcal{P}}}^t$ and extracts associated features $\bar{\bm{\mathcal{F}}}^s{=}\{\bar{\bm{f}}^s_i{\in}\mathbb{R}^{b}|i{=}1, 2, ..., \bar{N}_s\}$ and $\bar{\bm{\mathcal{F}}}^t{=}\{\bar{\bm{f}}^t_j{\in}\mathbb{R}^{b}|j{=}1, 2, ..., \bar{N}_t\}$, respectively. $b$ is dimension. 
Then, self- and cross-attention are applied to encode contextual information of two point clouds with partial overlaps, which outputs conditioned features $\bar{\bm{\mathcal{F}}}^s$ and $\bar{\bm{\mathcal{F}}}^t$.
Finally, the shared decoder starts with conditioned features $\bar{\bm{\mathcal{F}}}^s$ and $\bar{\bm{\mathcal{F}}}^t$, and outputs the point-wise feature descriptor $\bm{\mathcal{F}}^{s}{\in}\mathbb{R}^{N_s\times d}$ and $\bm{\mathcal{F}}^t{\in}\mathbb{R}^{N_t\times d}$ and overlap scores $\bm{O}^s{=}\{o_i^s\}{\in}\mathbb{R}_+^{N_s}$ and $\bm{O}^t{=}\{o_j^t\}{\in}\mathbb{R}_+^{N_t}$. 
$d$ is the dimension of features. The decoder combines NN-upsampling with linear layers and includes skip connections from the corresponding encoder layers.
For more details on feature extraction,  please refer to the supplementary material.

\vspace{0.1cm}
\noindent\textbf{Learning Posterior.}
Different from the previous works~\cite{yuan2020deepgmr,eckart2018hgmr} only considering the spatial coordinates of the points in the probabilistic registration model, we propose a method to learn the joint distribution over the spatial coordinate and feature spaces.
Specifically, we apply a multi-layer perceptron (MLP), i.e., cluster head $\psi$,  that takes as input $\bm{\mathcal{F}}^s$ and $\bm{\mathcal{F}}^t$ and outputs joint log probabilities and a Softmax operator that acts on log probabilities to generate probability matrices $\bm{S}^s{=}\{s^s_{ij}\}_{i,j=1}^{N^s,L{-}1}$ and $\bm{S}^t{=}\{s^t_{ij}\}_{i,j=1}^{N^t,L{-}1}$, respectively. 
To deal with outliers, it is straightforward to add a Gaussian kernel density. We define $\hat{\bm{S}}^x{=}\{s^x_{ij}\}_{i,j=1}^{N^x,L}~(x{\in}\{s,t\})$ with elements satisfying $\hat{s}^x_{iL}{=}1.0{-}o^x_{i}$ and $\hat{s}^x_{iL}{=}o^x_{i}s^x_{ij}, 1\leq j< L$.
\ourmethod assumes that coordinate and feature spaces share the same probability matrix (posterior distribution).
The GMM parameters $\bm{\Theta}^x$ for point cloud $\bm{\mathcal{P}}^x$, in 3D coordinate space, consists of $L$ triples $(\pi^x_{j},\bm{\mu}^x_{j},\bm{\Sigma}^x_{j})$, where $\pi^x_{j}$ is the mixing weight of component $j$ satisfying $\sum^L_{j=1}\pi^x_{j}=1$, $\bm{\mu}^x_{j}$ is a $3{\times} 1$ mean vector and $\bm{\Sigma}^x_{j}$ is a $3{\times} 3$ covariance matrix of the $j$-th component. 
Given the outputs $\bm{S}^x$ of $\psi$ together with the point coordinates $\bm{\mathcal{P}}^x$, the GMMs are calculated as:
\vspace{-0.2cm}
\begin{equation}\label{eq:gmm_3d}
\small
    \begin{aligned}
    &\pi^x_{j} {=} \frac{1}{N_x}\sum_{i=1}\hat{s}^x_{ij}, \bm{\mu}^x_{j} {=} \frac{1}{N_x\pi^x_{j}}\sum_{i=1}\hat{s}^x_{ij}\bm{p}^x_i, \\
    &\bm{\Sigma}^x_{j} {=} \sum_{i=1}\hat{s}^x_{ij}\left(\bm{p}^x_i {-} \bm{\mu}^x_{j}\right)\left(\bm{p}^x_i{-}\bm{\mu}^x_{j}\right)^\top, \\
    &G^{x}\left(\bm{x}\right) {=} \sum _{j=1}\pi^x_{j}\mathcal{N} \left(\bm{x}|\bm{\mu}^x_{j}, \bm{\Sigma}^x_{j}\right),x\in\{s, t\}.
    \end{aligned}
\vspace{-0.2cm}
\end{equation}
Similar in the coordinate space, based on probability matrices $\bm{S}^s$ and $\bm{S}^t$, the GMM parameters of point clouds $\bm{\mathcal{P}}^s$ and $\bm{\mathcal{P}}^t$ in feature space are also computed as:
\vspace{-0.2cm}
\begin{equation*}
\small
\bm{\mu}^{f_x}_j{=}\sum_{i=1}^{N_x}\frac{\hat{s}^x_{ij}\bm{f}^x_i}{N_x\pi^x_{j}}, \bm{\Sigma}^{f_x}_j {=} \sum_{i=1}^{N_x}\hat{s}^x_{ij}\left(\bm{f}^x_{i}{-}\bm{\mu}^{f_x}_j\right)\left(\bm{f}^x_{i}{-}\bm{\mu}^{f_x}_j\right)^\top,
\vspace{-0.2cm}
\end{equation*}
where subscript $x{\in}\{s, t\}$. Note that the GMMs in coordinate and feature spaces share mixing coefficients. For simplify, we denote $\Phi_k^{f_x}(\bm{x}){=}\mathcal{N} \left(\bm{x}|\bm{\mu}^{f_x}_{k}, \bm{\Sigma}^{f_x}_{k}\right)$ with $k{\in}\{1,\cdots, L\}$. The GMMs of point clouds $\bm{\mathcal{P}}^s$ and $\bm{\mathcal{P}}^t$ in feature space are then given as:
\vspace{-0.2cm}
\begin{equation} \label{eq:postf}
\small
    G^{f_s}\left(\bm{x}\right) {=} \sum_{j{=}1}^{L}\pi^s_{j}\Phi^{f_s}_j(\bm{x}), \quad
    G^{f_t}\left(\bm{x}\right) {=} \sum_{j{=}1}^{L}\pi^t_{j}\Phi^{f_t}_j(\bm{x}).
\vspace{-0.2cm}
\end{equation}

\noindent\textbf{Cluster-Level Matching.}
Instead of indirectly performing the maximum likelihood estimation between $G^s$ and $G^t$, weighted distribution-level correspondences are represented as soft assignments to the components based on the mixing weights of GMMs and the $L_2$ distance~\cite{jian2010robust} of distribution in the feature space. This is because $(\pi^s_{j},\bm{\mu}^s_{j},\bm{\Sigma}^s_{j})$ and $(\pi^t_{j},\bm{\mu}^t_{j},\bm{\Sigma}^t_{j})$ are not wholly matched when two point clouds are partially overlapped. Moreover, the aligned components should have similar mixing weights and small distances. To estimate the correspondences, we first calculate the distance between two GMMs as follows:
\begin{equation}
\small
\mathcal{D}(\Phi^{f_s}_i,\Phi^{f_t}_j)=\int_{\mathbb{R}}\left(\Phi^{f_s}_i(\bm{x})-\Phi^{f_t}_j(\bm{x})\right)^2d\bm{x}.
\end{equation}
We denote $r^x{=}\sum^{L-1}_{i=1}\max(\frac{\pi^x_i}{1-\pi^{x}_L}-\frac{\pi^y_i}{1-\pi^{y}_L},0)$ and cost matrix $\bm{D}$ with elements satisfying $\bm{D}_{ij}{=}\mathcal{D}(\Phi^{f_s}_i,\Phi^{f_t}_j)$. $x=s,y=t$ or $x=t,y=s$. 
In partially overlapping registration, some components are occluded in the other frame. Similar to~\cite{yu2021cofinet}, we propose here to solve it directly by changing the cost matrix as $\hat{\bm{D}}$ with elements satisfying,  if $i, j{<} L$, $\hat{\bm{D}}_{ij}{=}\bm{D}_{ij}$ otherwise $\bm{D}_{ij}{=}z$. $z$ is a learnable parameter.
The extended assignment matrix $\Gamma{\in}\mathbf{R}^{L\times L}$ can be estimated by solving the following optimization problem:
\vspace{-0.2cm}
\begin{equation}\label{eq:feature}
\begin{aligned}
& \min_{\Gamma}\sum_{ij}\Gamma_{ij}\hat{\bm{D}}_{ij},\\
& \mbox{s.t.,}~ \Gamma\bm{1}_{L}  {=} \hat{\bm{\pi}}^{s}, \Gamma^\top\bm{1}_{L} {=} \hat{\bm{\pi}}^{t}, \Gamma_{ij}\in[0, 1],
\end{aligned}
\vspace{-0.2cm}
\end{equation}
where $\hat{\bm{\pi}}^{x}{=}\frac{1}{1+r^x-\pi^{x}_L}(\pi^{x}_1, \pi^{x}_2, \cdots, \pi^{x}_{L{-}1}, r^x), x{\in}\{s, t\}$. We run the Sinkhorn
Algorithm~\cite{cuturi2013sinkhorn} to seek an optimal solution.
After that, each entry $(i,j)$ of $\Gamma$ implies the matching confidence between components. Following~\cite{yu2021cofinet}, we pick correspondences whose confidence scores are above a threshold $\tau=0.1$.
We define the picked distribution-level correspondence set as $\bar{C}{=}\{(\bar{\bm{\mu}}^s_i,\bar{\bm{\mu}}^t_i)\}$.

\vspace{0.1cm}
\noindent\textbf{Point-Level Registration.}
We first partition the points into clusters by assigning each point to its closest centroid in the geometric space. Once grouped, we obtain 3D patches comprised of points along with their corresponding clustering scores and descriptors. These patches enable us to extract point correspondences.
For a centroid $\bm{\mu}^s_i$, its associated point set $\mathrm{C}^s_{i}$ and feature set $\mathrm{F}^s_{i}$ are denoted as:
\vspace{-0.2cm}
\begin{equation*}
\small
	\begin{cases}
		\mathrm{C}^s_{i}=\{\bm{p}^s\in \bm{\mathcal{P}}^s\big|\|\bm{p}^s-\bm{\mu}^s_i\|_2\leq\|\bm{p}^s-\bar{\bm{\mu}}^s_j\|_2, i\neq j\}, \\
		\mathrm{F}^s_{i}=\{\bm{f}^s_{j}\in \bm{\mathcal{F}}^s\big|\bm{p}^s_j \in \mathrm{C}^s_{i}\},\\
		\mathrm{S}^s_{i}=\{\bm{s}^s_{ji}\in \bm{s}^s_i\big|\bm{p}^s_j \in \mathrm{C}^s_{i}\}.
	\end{cases}
 \vspace{-0.2cm}
\end{equation*}
The same operator is also performed for $\bm{\mu}^t_j$ and we get $\mathrm{C}^t_{i}$, $\mathrm{F}^t_{i}$, and $\mathrm{S}^t_{i}$. The cluster-level correspondence set $\mathcal{M}^\prime$ are expanded to its corresponding 3D patch, both in geometry space $\mathcal{M}_C{=}\{(\mathrm{C}^s_{i}, \mathrm{C}^t_{i})\}$, feature space $\mathcal{M}_F{=}\{(\mathrm{F}^s_{i}, \mathrm{F}^t_{i})\}$, and normalized clustering scores $\mathcal{M}_S{=}\{(\mathrm{S}^s_{i}, \mathrm{S}^t_{i})\}$. 
For computational efficiency, every patch samples the $K$ number of points based on the probability. Similar to cluster-level prediction, given a pair of overlapped patches $(\mathrm{C}^s_{i}, \mathrm{F}^s_{i}, \mathrm{S}^s_{i})$ and $(\mathrm{C}^t_{i}, \mathrm{F}^t_{i}, \mathrm{S}^t_{i})$, extracting point correspondences is  to match two smaller corresponded scale point clouds $(\mathrm{C}^s_{i}, \mathrm{C}^t_{i})$ by solving an optimization problem:
\vspace{-0.2cm}
\begin{equation}\label{eq:point}
 \min_{\bm{\Gamma}^i}\left<\bm{D}^i, \bm{\Gamma}^i\right>,
 \mbox{s.t.,~}{\bm{\Gamma}^i}^\top\bm{1}_K=\mathrm{S}^t_{i}, \bm{\Gamma}^i\bm{1}_K=\mathrm{S}^s_{i},
 \vspace{-0.2cm}
\end{equation}
where each $\bm{\Gamma}^i{=}[\bm{\Gamma}^i]^{K\times K}_{kl}$ represents an assignment matrix and $\bm{D}^i{=}[\bm{D}^i]_{kl}$ with $\bm{D}^i_{kl}{=}\|\frac{\mathrm{F}^s_{i}\left(k\right)}{\|\mathrm{F}^s_{i}(k)\|_2}{-}\frac{\mathrm{F}^t_{i}(l)}{\|\mathrm{F}^t_{i}(l)\|_2}\|_2$. 
After reaching $\bm{\Gamma}^i$, we select correspondences from $(\mathrm{C}^s_{i}, \mathrm{C}^t_{i})$ with maximum confidence score for each row of $\bm{\Gamma}^i$. 
We denote each correspondence set extracted from a pair of patches as $\mathcal{M}^i{=}\{(\bm{p}^s_{\hat{i}} {\in} \mathrm{C}^s_{i},\bm{p}^t_{\hat{j}} {\in} \mathrm{C}^t_{i}),\hat{i}{=}1,2,\cdots,K\big|\hat{j} {=} \arg \max_{k} \bm{\Gamma}^{i}_{\hat{i},k}\}$.
The final point correspondence set $\mathcal{M}$ consists of the union of all the obtained patch-level correspondence sets $\mathcal{M}^i$. Following \cite{bai2021pointdsc,yu2021cofinet}, a variant of RANSAC \cite{fischler1981random} that is specialized to 3D registration takes $\mathcal{M}$ as an input to estimate the transformation.

\subsection{Consistency-Based Unsupervised Learning}
\noindent\textbf{Self-Consistency Loss.}
Our self-consistency loss encourages point clouds to share an identical posterior distribution in coordinate and feature spaces. It can be directly used without using any data augmentation.
Because training the network parameters is equivalent to optimizing the $\bm{\Theta}^s$ and $\bm{\Theta}^t$, the GMMs parameters can be fitted to the observed data points via maximizing the log-likelihood of samples to $\bm{\Theta}^s$ and $\bm{\Theta}^t$.
However, the log-likelihood function is unstable in the training processing since its value goes to infinity for a specific combination of means and some degenerating covariance matrices.
To avoid covariance degeneration, we approximate the probabilities of points belonging to each cluster based on their distance to the centroids estimated by Eq.~\eqref{eq:gmm_3d} under the constraints of the mixture weights.  
We denote the empirical distribution matrices of $\bm{\mathcal{P}}^s$ and $\bm{\mathcal{P}}^t$ as $\bm{\gamma}^s=\{\bm{\gamma}^s_{ij}\}$ and $\bm{\gamma}^t=\{\bm{\gamma}^t_{ij}\}$.
This results in the following optimization objective:
\vspace{-.2cm}
\begin{equation}\label{eq:sgamma}
\begin{aligned}
& \min_{\bm{\gamma}^x}\sum_{i,j}\bm{\gamma}^x_{ij}\|\bm{p}^x_i-\bm{\mu}^x_j\|^2_2,\\
& \mbox{s.t.,}~ \sum_i\bm{\gamma}^x_{ij} {=} N_x\bm{\pi}^{x}_j, \sum_j\bm{\gamma}^x_{ij} {=} 1, \bm{\gamma}_{ij}\in[0, 1],
\end{aligned}
\vspace{-.2cm}
\end{equation}
where $x \in \{s, t\}$. 
$\sum_j\bm{\gamma}^x_{ij}{=}1$ is based on the property of the probability that the sum of all the probabilities for all possible events is equal to one. 
$\sum_i\bm{\gamma}^x_{ij} {=} N_x\bm{\pi}^{x}_j$ is the constraints of the mixture weights.
We address the minimization of Eq.~\eqref{eq:sgamma} by adopting an efficient version of the Sinkhorn algorithm~\cite{cuturi2013sinkhorn}. Coordinate and feature spaces share an identical posterior distribution means that $\bm{S}^x$ and $\bm{\gamma}^x$ should be equal, which leads to a cross-entropy loss. Our self-consistency loss is thus formulated as follows:
\vspace{-0.2cm}
\begin{equation}\label{eq:sc}
    \mathcal{L}_{sc} = -\sum_{ij}\bm{\gamma}^s_{ij}\log s^t_{ij} - \sum_{ij}\bm{\gamma}^t_{ij}\log s^t_{ij}.
\vspace{-0.2cm}
\end{equation}

\noindent\textbf{Cross-Consistency Loss.} 
The described self-consistency loss only encourages the learned representation to be spatially sensitive, but it cannot ensure that the learned features be transformation invariant.
Therefore, we introduce a cross-consistency loss to encourage the network to learn transformation-invariant feature representations.
Our cross-consistency loss is based on the fact that
the cluster labeling should not change if the points are rigidly transformed.
This fact means that if points $\bm{p}^s{\in}\bm{\mathcal{P}}^s$ and $\bm{p}^t{\in}\bm{\mathcal{P}}^t$ belong to the same cluster, they should share the same cluster centroid.
Therefore, the cross-consistency loss can make full use of the information from both aligned point clouds.
Concretely, for two input features sets $\left(\bm{\mathcal{F}}^s, \bm{\mathcal{F}}^t\right)$, and two probability matrices $\left(\bm{S}^s, \bm{S}^t\right)$, we obtain a new feature set $\bm{\mathcal{F}} {=} cat\left(\bm{\mathcal{F}}^s, \bm{\mathcal{F}}^t\right)$ and a probability matrix $\bm{\mathcal{S}} {=} cat\left(\bm{S}^s, \bm{S}^t\right)$. 
$cat(\cdot,\cdot)$ means concatenation.
We assume the current estimated rotation and translation are $\bm{R}$ and $\bm{t}$. We define $\bm{\mathcal{\bar{P}}}^s{=}\bm{R}\bm{\mathcal{P}}^s{+}\bm{t}$  and $\bm{\mathcal{P}}{=}cat(\bm{\mathcal{\bar{P}}}^s,\bm{\mathcal{P}}^t)$.
Then, we calculate the parameters of global GMMs in both feature and euclidean spaces as:
\vspace{-0.2cm}
\begin{equation*}
\small
   \pi_j = \frac{\sum_{i}\bm{s}_{ij}}{N},
    ~\bm{\mu}^{f}_j = \frac{\sum_{i}\bm{s}_{ij}\bm{f}_{i}}{\pi_jN}, ~\bm{\mu}^{e}_j = \frac{\sum_{i}\bm{s}_{ij}\bm{p}_{i}}{\pi_jN},
\vspace{-0.2cm}
\end{equation*}
where $N{=}N_s{+}N_t$.
To avoid two aligned point clouds being grouped into separate clusters, we assume that clustering satisfies two constraints:
\begin{itemize} [leftmargin=*]
\item GMMs are coupled with approximate uniform mixing weights in coordinate and feature spaces.
\item If a point $\bm{p}_i$ belongs to partition $j$, point $\bm{p}_i$ and its coupled centroid should have the shortest distance.
\end{itemize}
Let $\bm{\gamma}{=}\{\gamma_{ij}\}$ to be the empirical probability matrix. 
The two constraints can then be ensured by minimizing the following objective:
\vspace{-.2cm}
\begin{equation}\label{eq:mgamma}
\small
    \begin{aligned}
        & \min_{\bm{\gamma}}\sum_{ij}\left(\lambda_1\|\bm{p}_i-\bm{\mu}^e_j\|^2_2+\lambda_2\|\bm{f}_i-\bm{\mu}^f_j\|^2_2\right)\gamma_{ij}, \\
        & \mbox{s.t.,} \sum_i\gamma_{ij} = 1, \sum_j\gamma_{ij} = \frac{N}{L}, \gamma_{ij}\in[0, 1],
    \end{aligned}
\vspace{-.2cm}
\end{equation}
where $\lambda_i {\in} [0,1]$ are learned parameters. 
After solving Eq.~\eqref{eq:mgamma}, we then infer our cross-consistency loss as:
\vspace{-.2cm}
\begin{equation}\label{eq:scsup}
    \mathcal{L}_{cc}(\bm{\gamma}, \bm{S}) = -\sum_{ij}\bm{\gamma}_{ij}\log s_{ij},
\vspace{-.2cm}
\end{equation}
which corresponds to the minimization of the standard cross-entropy loss between $\bm{\gamma}$ and predictions $\bm{S}$. 

\vspace{0.1cm}
\noindent\textbf{Local Contrastive Loss.}
The local neighbors provide essential information for feature learning on the objects of the point clouds~\cite{li2022hybridcr}. 
For instance, occlusions and holes always occur in objects in indoor and outdoor scenes~\cite{li2022hybridcr}. 
If the network captures the local structure information from other complete objects, it can boost the model robustness on incomplete objects during training. 
While the local descriptors of the point clouds mainly derive from the points and their neighbors~\cite{li2022hybridcr}, which motivates us to model the local information of the point cloud by introducing local contrastive loss. 
Specifically, given a centroid $\bm{\mu}^x_i$ of point cloud $\bm{\mathcal{P}}^x$ with $x{\in}\{s,t\}$, we search its nearest point $\bm{p}^x_i$ and associated feature vector $\bm{f}^x_i$ by the point-wise Euclidean distance. Based on this, we construct the local contrastive loss $\mathcal{L}_{lc}$ following InfoNCE~\cite{xie2020pointcontrast} by pulling $\bm{f}^x_i$ close to $\bm{\mu}^x_i$, while pushing it away from the neighbor vector of other points. We also encourage $\bm{\mu}^{f_s}_i$ and $\bm{\mu}^{f_t}_i$ to be similar:
\vspace{-0.2cm}
\begin{equation*}
\small
\begin{aligned}
&\mathcal{L}_{lc} {=}{-}\frac{1}{L}\sum^L_{i=1}
\log \frac{\exp\left(\bm{\mu}^{f_s}_i{\bm{\mu}^{f_t}_i}^\top\right)}{\sum^L_{j=1}\exp\left(\bm{\mu}^{f_s}_i{\bm{\mu}^{f_t}_j}^\top\right)}{-}\\
&\frac{1}{L}\sum^L_{i=1}\log \frac{\exp\left(\bm{\mu}^{f_s}_i{\bm{f}^s_i}^\top\right)\exp\left(\bm{\mu}^{f_t}_i{\bm{f}^t_i}^\top\right)}{\sum^L_{j=1}\exp\left(\bm{\mu}^{f_s}_i{\bm{f}^s_j}^\top\right)\sum^L_{j=1}\exp\left(\bm{\mu}^{f_t}_i{\bm{f}^t_j}^\top\right)}.
\end{aligned}
\vspace{-0.2cm}
\end{equation*}
Thus, the final loss is the combination of self-consistency loss, cross-consistency loss, and local contrastive loss as:
\vspace{-0.2cm}
\begin{equation}
\small
    \mathcal{L} = \mathcal{L}_{sc} + \mathcal{L}_{cc} + \mathcal{L}_{lc}.
    \vspace{-0.2cm}
\end{equation}
In particular, different from most existing methods, the correspondence or pose between two partially overlapping point clouds is unknown in our training processing.


\section{Experiments}\label{sec:exp}
We conduct extensive experiments to evaluate the performance of our method on the real datasets 3DMatch \cite{zeng20173dmatch} and 3DLoMatch~\cite{huang2021predator}, as well as on the synthetic datasets ModelNet~\cite{wu20153d} and ModelLoNet~\cite{huang2021predator}.

\subsection{Implementation Details} 
Our method is implemented in PyTorch and was trained on one Quadro GV100 GPU (32G) and two Intel(R) Xeon(R) Gold 6226 CPUs.  
We used the AdamW optimizer with an initial learning rate of $1e{-}4$ and a weight decay of $1e{-}6$.  
We adopted the similar encoder and decoder architectures used in~\cite{qin2022geometric}.
For the 3DMatch dataset, we trained for 200 epochs with a batch size of 1, halving the learning rate every 70 epochs.
We trained on the ModelNet for 400 epochs with a batch size of 1, halving the learning rate every 100 epochs. 
On 3DMatch and 3DLoMatch, we set $L{=}128$ with truncated patch size $K{=}64$. 
On ModelNet and ModelLoNet, we set $L{=}64$ with truncated patch size $K{=}32$. 
The cluster head MLP consists of 3 fully connected layers. Each layer is composed of a linear layer followed by batch normalization. The hidden layer and the final linear layer output dimension are 512 and clusters, respectively. Except for the final layer, each layer has a LeakyReLU activation.

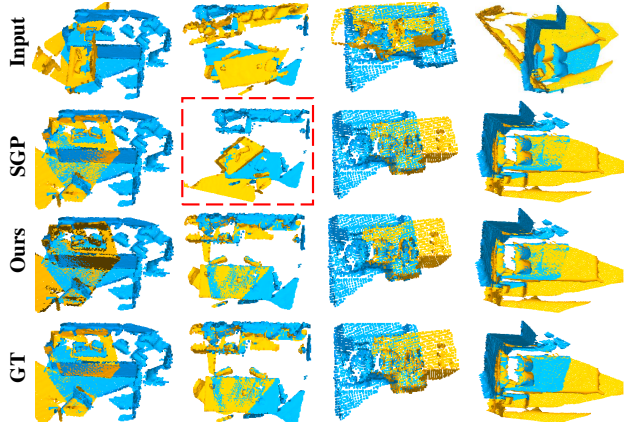
\begin{figure}[t]
	\centering 
\begin{overpic}[width=0.95\columnwidth]{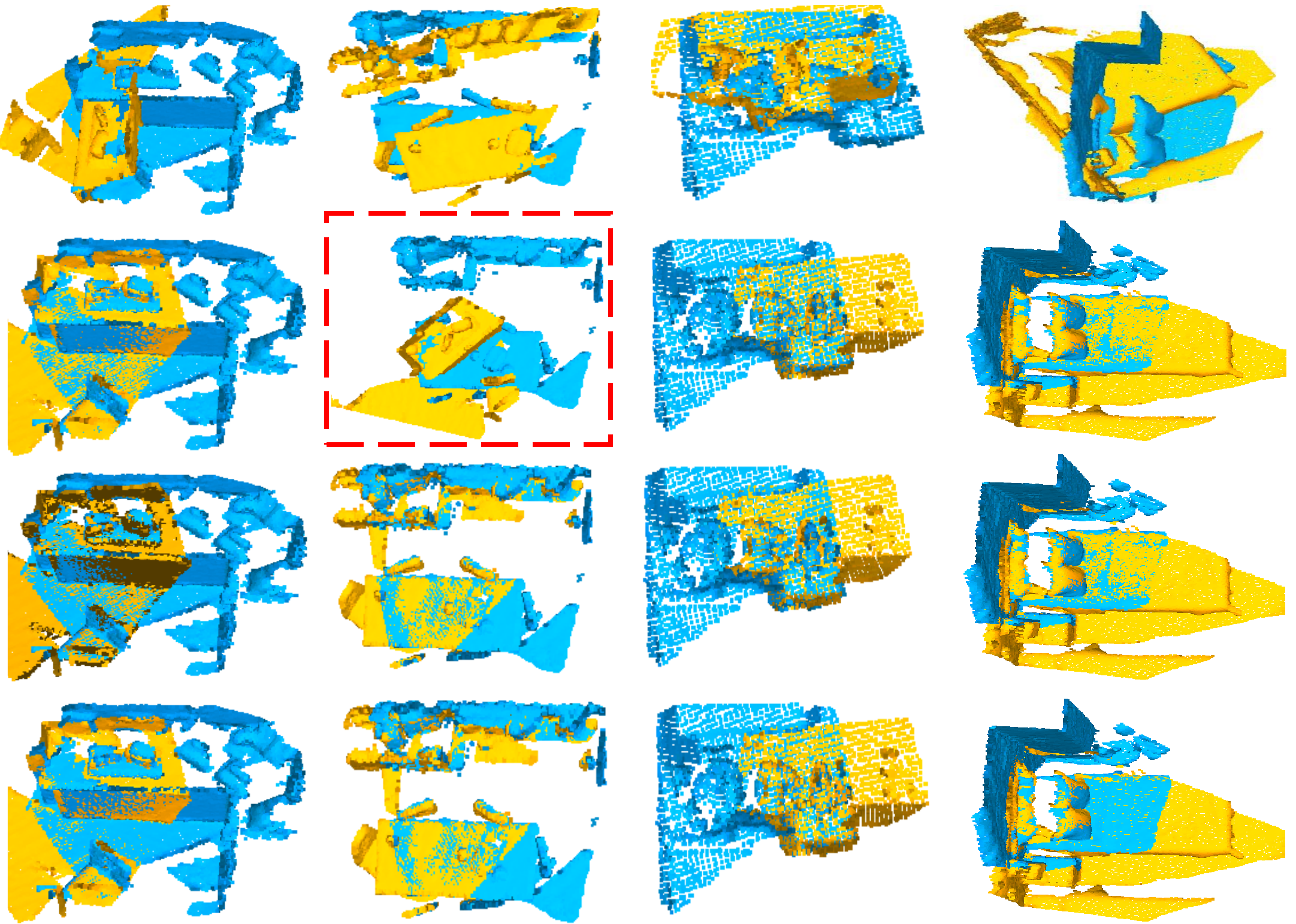}
    \put(-3.7,60.0){\color{black}\footnotesize\rotatebox{90}{\textbf{Input}}}
    \put(-3.7,41.5){\color{black}\footnotesize\rotatebox{90}{\textbf{SGP}}}
    \put(-3.7,25.4){\color{black}\footnotesize\rotatebox{90}{\textbf{Ours}}}
    \put(-3.7,7.0){\color{black}\footnotesize\rotatebox{90}{\textbf{GT}}}
\end{overpic}
\caption{Example qualitative registration results for 3DMatch. The unsuccessful cases are enclosed in red boxes.}
	\label{fig:3dvs}
 \vspace{-0.4cm}
\end{figure}

\subsection{Evaluation on 3DMatch and 3DLoMatch}
\noindent\textbf{Datasets and Metrics.} 3DMatch \cite{zeng20173dmatch} and 3DLoMatch \cite{huang2021predator} are two widely used indoor datasets with more than $30\%$ and $10\% {\sim} 30\%$ partially overlapping scene pairs, respectively. 3DMatch contains 62 scenes, from which we use 46 for training, 8 for validation, and 8 for testing. The test set contains 1,623 partially overlapping point cloud fragments and their corresponding transformation matrices. We used training data preprocessed by \cite{huang2021predator} and evaluated with both the 3DMatch and 3DLoMatch protocols.  Each input point cloud contains an average of about 20,000 points. We performed training data augmentation by applying small rigid perturbations, jittering the point locations, and shuffling points.
Following REGTR~\cite{yew2022regtr} and SGP~\cite{yang2021self}, we evaluated the Relative Rotation Errors (RRE) and Relative Translation Errors (RTE) that measure the accuracy of successful registrations. We also assessed Registration Recall (RR), the fraction of point cloud pairs whose transformation error is smaller than a threshold (i.e., 0.2m).

\begin{table}[!tbp]
	\centering
	\caption{Results on both 3DMatch and 3DLoMatch datasets. The best results for each criterion are labeled in bold, and the best results of unsupervised methods are underlined.}
	\resizebox{1\linewidth}{!}{%
	\begin{tabular}{r | c c c | c c c}
	\toprule
	~ &  \multicolumn{3}{c|}{3DMatch} &  \multicolumn{3}{c}{3DLoMatch} \\
	Method  & RR$\uparrow$ & RRE $\downarrow$ & RTE $\downarrow$
    & RR $\uparrow$ & RRE $\downarrow$ & RTE $\downarrow$ \\
    \midrule
    & \multicolumn{6}{c}{Supervised Methods} \\
    \midrule
	FCGF\cite{choy2019fully}  
    & 85.1\% & 1.949 & 0.066 
    & 40.1\% & 3.147 & 0.100 \\
	D3Feat\cite{bai2020d3feat} 	
    & 81.6\% & 2.161 & 0.067 
    & 37.2\% & 3.361 & 0.103 \\
    OMNet~\cite{xu2021omnet} 
    & 35.9\% & 4.166 & 0.105 
    & 8.4\%  & 7.299 & 0.151 \\
	DGR~\cite{choy2020deep} 
    & 85.3\% & 2.103 & 0.067 
    & 48.7\% & 3.954 & 0.113 \\
	Predator1K~\cite{huang2021predator} 
    & 90.5\% & 2.062 & 0.068 
    & 62.5\% & 3.159 & 0.096 \\
	CoFiNet\cite{yu2021cofinet}	  
    & 89.7\% & 2.147 & 0.067
    & 67.2\% & 3.271 & 0.090 \\
	GeoTrans~\cite{qin2022geometric} 
    & \bf92.0\% & 1.808 & 0.063 
    & \bf74.0\% & 2.934 & 0.089 \\
    REGTR~\cite{yew2022regtr} 
    & \bf92.0\% & \bf1.567 & \bf0.049
    & 64.8\% & \bf2.827 & \bf0.077 \\
    \midrule
    & \multicolumn{6}{c}{Unsupervised Methods} \\
    \midrule
    PPFFoldNet~\cite{deng2018ppf} 
    & 69.3\% & 3.021 & 0.089
    & 24.8\% & 7.527 & 1.884\\
    SGP + R10K~\cite{yang2021self}	  
    & 85.5\% & 1.986 & 0.079 
    & 39.4\% & 3.529 & 0.099\\
	\ourmethod (Ours)	
    & \underline{91.4\%} & \underline{1.642} & \underline{0.064}
    & \underline{64.3\%} & \underline{2.951} & \underline{0.086} \\			
	\bottomrule
	\end{tabular}
	}
\label{table:3dm}
	\vspace{-0.4cm}
\end{table}

\vspace{0.1cm}
\noindent\textbf{Baselines.} We chose supervised state-of-the-art (SOTA) methods: OMNet~\cite{xu2021omnet}, FCGF \cite{choy2019fully}, D3Feat \cite{bai2020d3feat}, SpinNet \cite{ao2021spinnet}, Predator \cite{huang2021predator}, REGTR~\cite{yew2022regtr}, CoFiNet \cite{yu2021cofinet}, and GeoTransformer\cite{qin2022geometric}, as well as unsupervised PPFFoldNet~\cite{deng2018ppf} and SGP~\cite{yang2021self} as our baselines.

\begin{figure}[t]
	\centering 
\begin{overpic}[width=0.95\columnwidth]{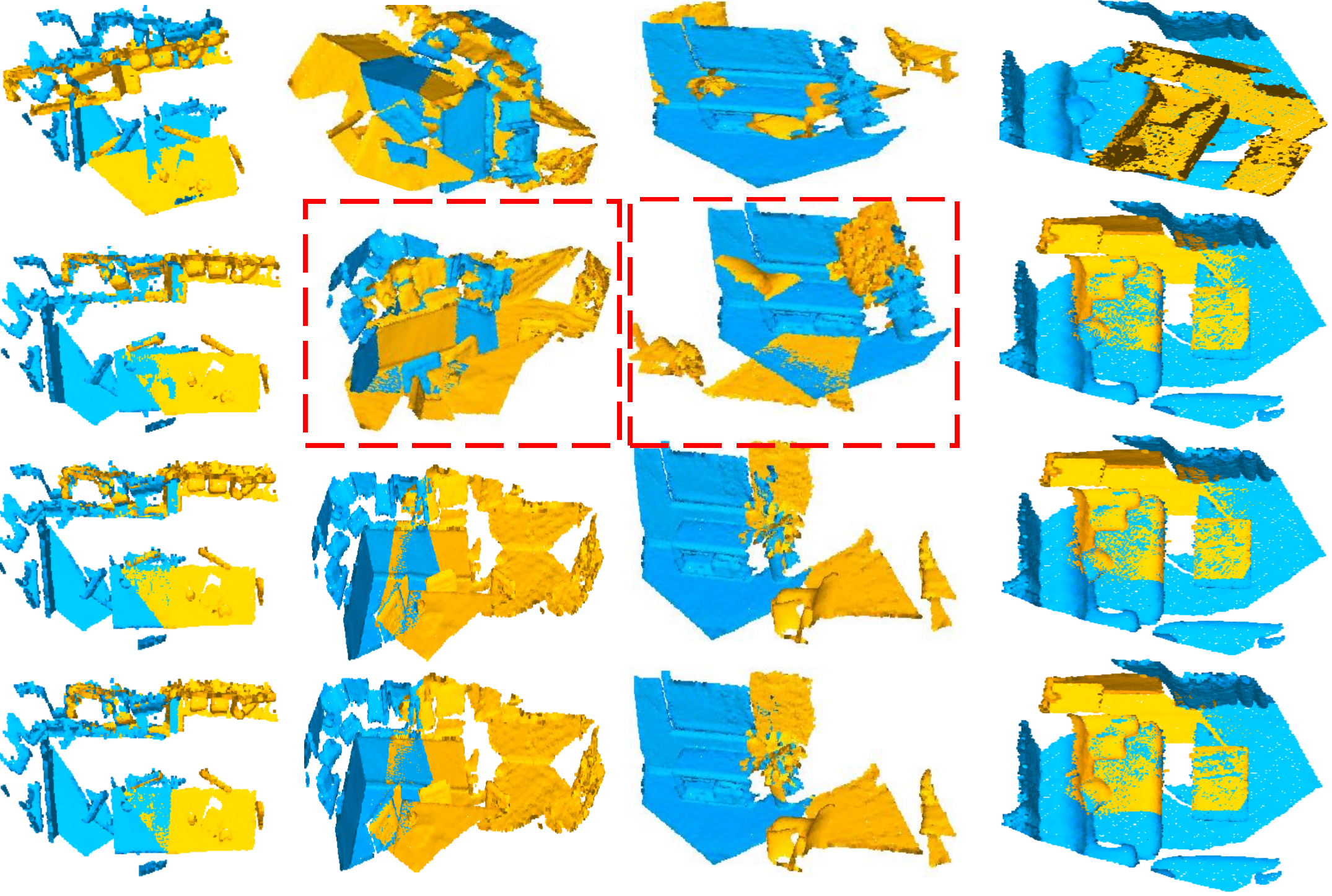}
    \put(-3.8,55.0){\color{black}\footnotesize\rotatebox{90}{\textbf{Input}}}
    \put(-3.8,40.0){\color{black}\footnotesize\rotatebox{90}{\textbf{SGP}}}
    \put(-3.8,24.0){\color{black}\footnotesize\rotatebox{90}{\textbf{Ours}}}
    \put(-3.8,7.0){\color{black}\footnotesize\rotatebox{90}{\textbf{GT}}}
\end{overpic}
\caption{Example qualitative registration results for 3DLoMatch. The unsuccessful cases are enclosed in red boxes.} \label{fig:3dvslo}
\vspace{-0.4cm}
\end{figure}

\vspace{0.1cm}
\noindent\textbf{Registration Results.} 
The results of various methods are shown in Table~\ref{table:3dm}, where the best performance is highlighted in bold while the best-unsupervised results are marked with an underline. For both 3DMatch and 3DLoMatch, our method outperforms all unsupervised methods and achieves the lowest average rotation (RRE) and translation (RTE) errors across scenes. Our method also achieves the highest average registration recall, which reflects the final performance on point cloud registration (91.4\% on 3DMatch and 64.3\% on 3DLoMatch). Specifically, \ourmethod largely exceeds the previous winner and our closest competitor, SGP,  (85.5\% RR on 3DMatch) by about 5.9\% and (39.4\% RR on 3DLoMatch) by 24.9\%. Interestingly, our method also exceeds some supervised methods, {\em e.g.} OGMM, FCGF, D3Feat, DGR, and Predator1K, showing its efficacy in both high- and low-overlap scenarios. Even compared with recent supervised SOTA methods, our method achieves competitive results. Figs.~\ref{fig:3dvs} and~\ref{fig:3dvslo} show examples of qualitative results on both 3DMatch and 3DLoMatch. GT indicates ground truth. SGP failed in one case of Fig.~\ref{fig:3dvs} on 3DMatch and failed in two cases of Fig.~\ref{fig:3dvslo} on 3DLoMatch, but our method succeeded in all cases. This is because our unsupervised method can learn more discriminative features and our matching strategy can deal with partial overlap registration, which further shows the effectiveness of \ourmethod.

\subsection{Evaluation on ModelNet40}
\noindent\textbf{Datasets and Metrics.} ModelNet40 \cite{wu20153d} contains 12,311 meshed CAD models from 40 categories. Following the data setup in~\cite{huang2021predator,yew2022regtr}, each point cloud is sampled from ModelNet40 with 1,024 points followed by cropping and sub-sampling into two partial overlap settings: ModelNet has 73.5\% pairwise overlap on average, and ModelLoNet contains a lower 53.6\% average overlap. We train only on ModelNet and generalize to ModelLoNet. We follow~\cite{yew2022regtr} and measure the performance using Relative Rotation Error (RRE) and Relative Translation Error (RTE) on all point clouds and as Chamfer distance (CD) between scans.

\begin{table}[!tbp]
	\centering
	\caption{Results on both ModelNet and ModelLoNet datasets. The best results for each criterion are labeled in bold, and the best results of unsupervised methods are underlined.}
	\resizebox{1\linewidth}{!}{%
	\begin{tabular}{r | c c c | c c c}
	\toprule
	~ &  \multicolumn{3}{c|}{ModelNet} &  \multicolumn{3}{c}{ModelLoNet} \\
	Method  & RRE $\downarrow$ & RTE $\downarrow$ & CD $\downarrow$
    & RRE $\downarrow$ & RTE $\downarrow$  & CD $\downarrow$ \\
    \midrule
    & \multicolumn{6}{c}{Traditional Methods} \\
    \midrule
    ICP~\cite{besl1992method} 
    & 13.74 & 0.132 & 0.1225
    & 24.13 & 0.224 & 0.1289\\
    FGR~\cite{zhou2016fast}
    & 28.68 & 0.160 & 0.1290
    & 34.39 & 0.244 & 0.1339\\
    CPD~\cite{myronenko2010point} 
    & 14.17 & 0.139 & 0.1277
    & 28.78 & 0.253 & 0.1320\\
    GMMReg~\cite{jian2010robust} 
    & 16.41 & 0.163 & 0.1304
    & 24.03 & 0.243 & 0.1298\\
    SVR~\cite{campbell2015adaptive}  
    & 14.40 & 0.140 & 0.1279
    & 23.45 & 0.222 & 0.1322\\
    FilterReg~\cite{gao2019filterreg} 
    & 24.07 & 0.193 & 0.1336
    & 37.28 & 0.298 & 0.1367\\
	\midrule
    & \multicolumn{6}{c}{Supervised Methods} \\
    \midrule
    DCP-v2~\cite{wang2019deep} 
    & 11.98 & 0.171 & 0.0117 
    & 16.50 & 0.300 & 0.0268 \\
    DeepGMR \cite{yuan2020deepgmr}
    & 7.871 & 0.108 & 0.0056
    & 9.867 & 0.117 & 0.0064\\
	OMNet~\cite{xu2021omnet} 
    & 2.947 & 0.032 & 0.0015 
    & 6.517 & 0.129 & 0.0074 \\
	RPM-Net~\cite{yew2020rpm} 
    & 1.712 & 0.018 & 0.0009
    & 7.342 & 0.124 & 0.0050 \\
	Predator~\cite{huang2021predator} 
    & 1.739 & 0.019 & 0.0009
    & 5.235 & 0.132 & 0.0083 \\
	GeoTrans~\cite{qin2022geometric} 
    & 2.145 & 0.020 & \bf0.0003
    & 4.741 & 0.103 & 0.0143 \\
    REGTR~\cite{yew2022regtr} 
    & 1.473 & 0.014 & 0.0008 
    & 3.930 & 0.087 & \bf0.0037 \\
    \midrule
    & \multicolumn{6}{c}{Unsupervised Methods} \\
    \midrule
    CEMNet~\cite{jiang2021sampling}
    & 2.575 & 0.019 & 0.0368
    & 9.417 & 0.151 & 0.0861\\
    RIENet \cite{shen2022reliable}
    & 2.447 & 0.018 & 0.0365
    & 14.49 & 0.105 & 0.0828\\
    UGMM~\cite{huang2022unsupervised}
    & 13.65 & 0.124 & 0.0753
    & 17.39 & 0.161 & 0.0745 \\
	\ourmethod (Ours)	
    & \bf\underline{1.331} & \bf\underline{0.011} & \underline{0.0306}
    & \bf\underline{3.578} & \bf\underline{0.069} & \underline{0.0416} \\			
	\bottomrule
	\end{tabular}
	}
\label{table:mnet}
\vspace{-0.4cm}
\end{table}

\begin{table}[!t]
	\centering
	\caption{The results of different combinations of loss functions in both ModelNet and ModelLoNet datasets. The best results for each criterion are labeled in bold.}
	\resizebox{0.9\linewidth}{!}{%
	\begin{tabular}{r | c c c | c c c}
	\toprule
	~ &  \multicolumn{3}{c|}{ModelNet} &  \multicolumn{3}{c}{ModelLoNet} \\
	Method  & RRE $\downarrow$ & RTE $\downarrow$ & CD $\downarrow$
    & RRE $\downarrow$ & RTE $\downarrow$  & CD $\downarrow$ \\
	\midrule   
    CC
    & 6.985 & 0.087 & 0.0357 
    & 8.176 & 0.084 & 0.0483\\
    SC
    & 5.898 & 0.045 & 0.0314
    & 8.104 & 0.081 & 0.0470\\
    LC
    & 7.871 & 0.046 & 0.0393
    & 8.790 & 0.091 & 0.0482\\
    SC + LC
    & 3.742 & 0.062 & 0.0324
    & 5.835 & 0.084 & 0.0334\\    
    CC + LC
    & 3.867 & 0.059 & 0.0314
    & 5.256 & \bf0.061 & 0.0422\\ 
    CC + SC  
    & 3.421 & 0.048 & 0.0360
    & 5.229 & 0.064 & 0.0423 \\
    CC + SC + LC 
    & \bf 1.331 & \bf 0.011 & \bf 0.0306
    & \bf 3.578 & 0.069 & \bf 0.0416 \\	 		
	\bottomrule
	\end{tabular}}
\label{tb:loss}
\vspace{-0.4cm}
\end{table}

\noindent\textbf{Baselines.} We chose recent supervised SOTA methods: DCP-v2~\cite{wang2019deep}, OMNet~\cite{xu2021omnet}, RPM-Net~\cite{yew2020rpm}, Predator \cite{huang2021predator}, REGTR~\cite{yew2022regtr}, CoFiNet \cite{yu2021cofinet}, and GeoTransformer\cite{qin2022geometric}, as well as unsupervised method RIENet~\cite{shen2022reliable} and UGMM~\cite{huang2022unsupervised} as our baselines.
For traditional methods, we choose point-level methods ICP~\cite{besl1992method} and FGR~\cite{zhou2016fast}, as well as probabilistic methods CPD~\cite{myronenko2010point}, GMMReg~\cite{jian2010robust}, SVR~\cite{campbell2015adaptive}, and FilterReg~\cite{gao2019filterreg} as baselines.
For Predator, RPM-Net, OMNet, and REGTR, we use the results provided in REGTR. In REGTR, Predator samples 450 points in the experiment, and OMNet obtained a slightly improved result in all categories.
We utilize the codes provided by the authors for probabilistic methods. To improve the results for partial registration, we replace PointNet with DGCNN in DeepGMR. Additionally, we use Open3D for ICP and FGR.

\begin{figure}[!t]
	\centering
    \begin{overpic}[width=0.95\columnwidth]{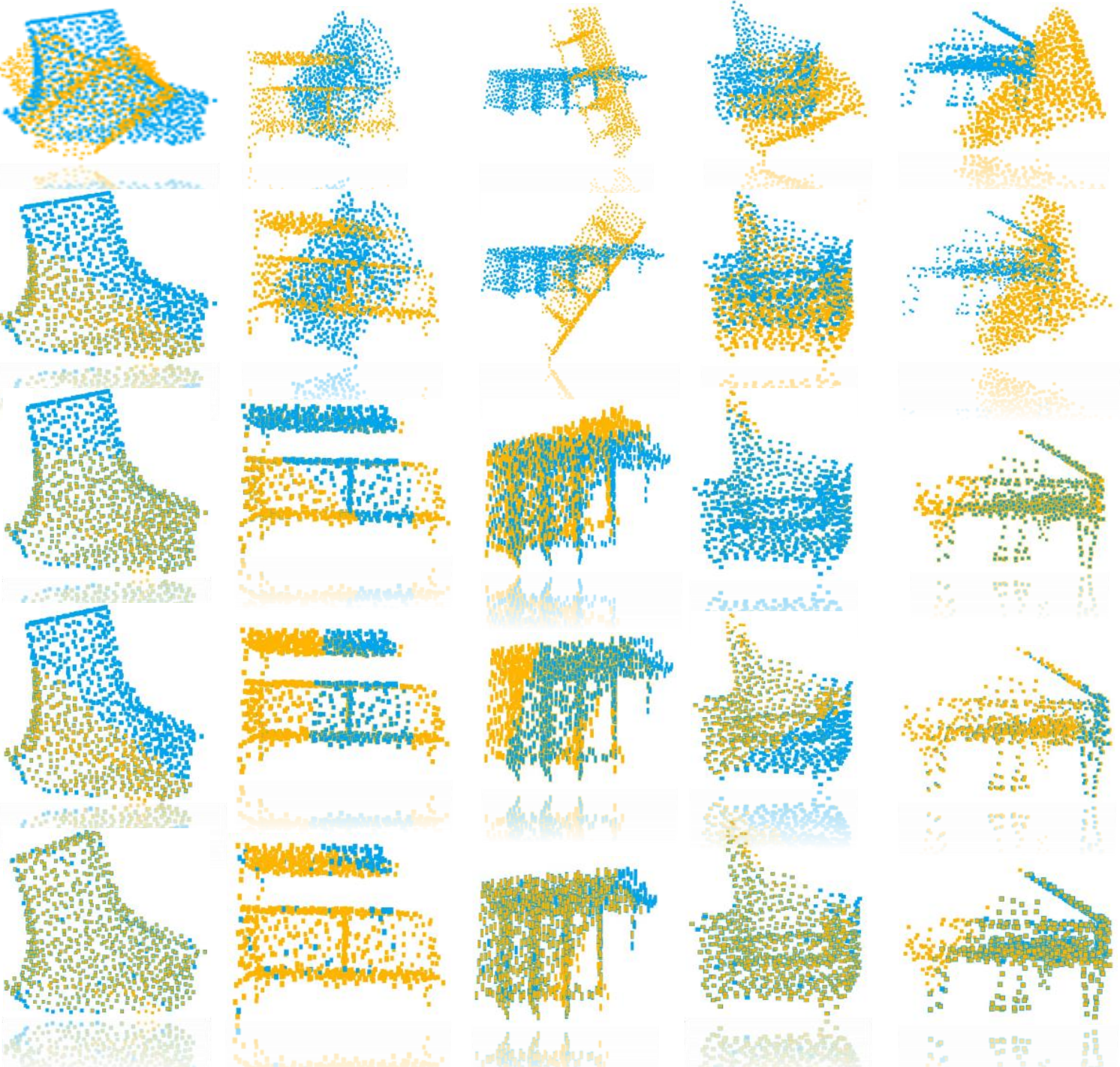}
    \put(-3.5,85.0){\color{black}\footnotesize\rotatebox{90}{\textbf{Input}}}
    \put(-3.5,69.0){\color{black}\footnotesize\rotatebox{90}{\textbf{ICP}}}
    \put(-3.5,48.0){\color{black}\footnotesize\rotatebox{90}{\textbf{RIENet}}}
    \put(-3.5,25.4){\color{black}\footnotesize\rotatebox{90}{\textbf{REGTR}}}
    \put(-3.5,10.0){\color{black}\footnotesize\rotatebox{90}{\textbf{Ours}}}
    \end{overpic}
	\caption{Registration results of different methods on ModelNet with more than 70\% partial overlaps.}
	\label{fig:mndet}
 \vspace{-0.3cm}
\end{figure}

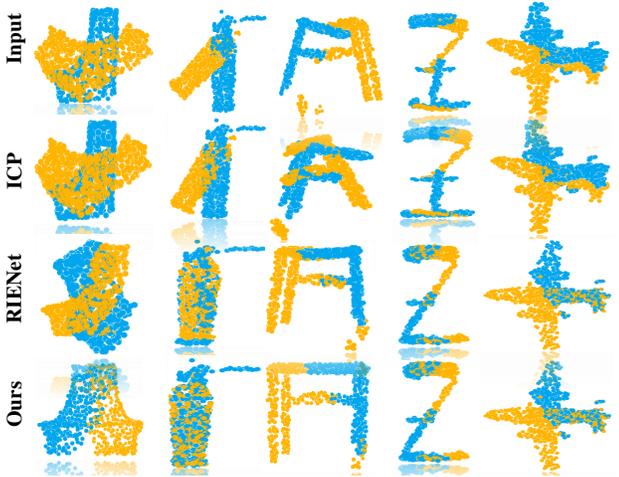
\begin{figure}[!t]
	\centering
    \begin{overpic}[width=0.95\columnwidth]{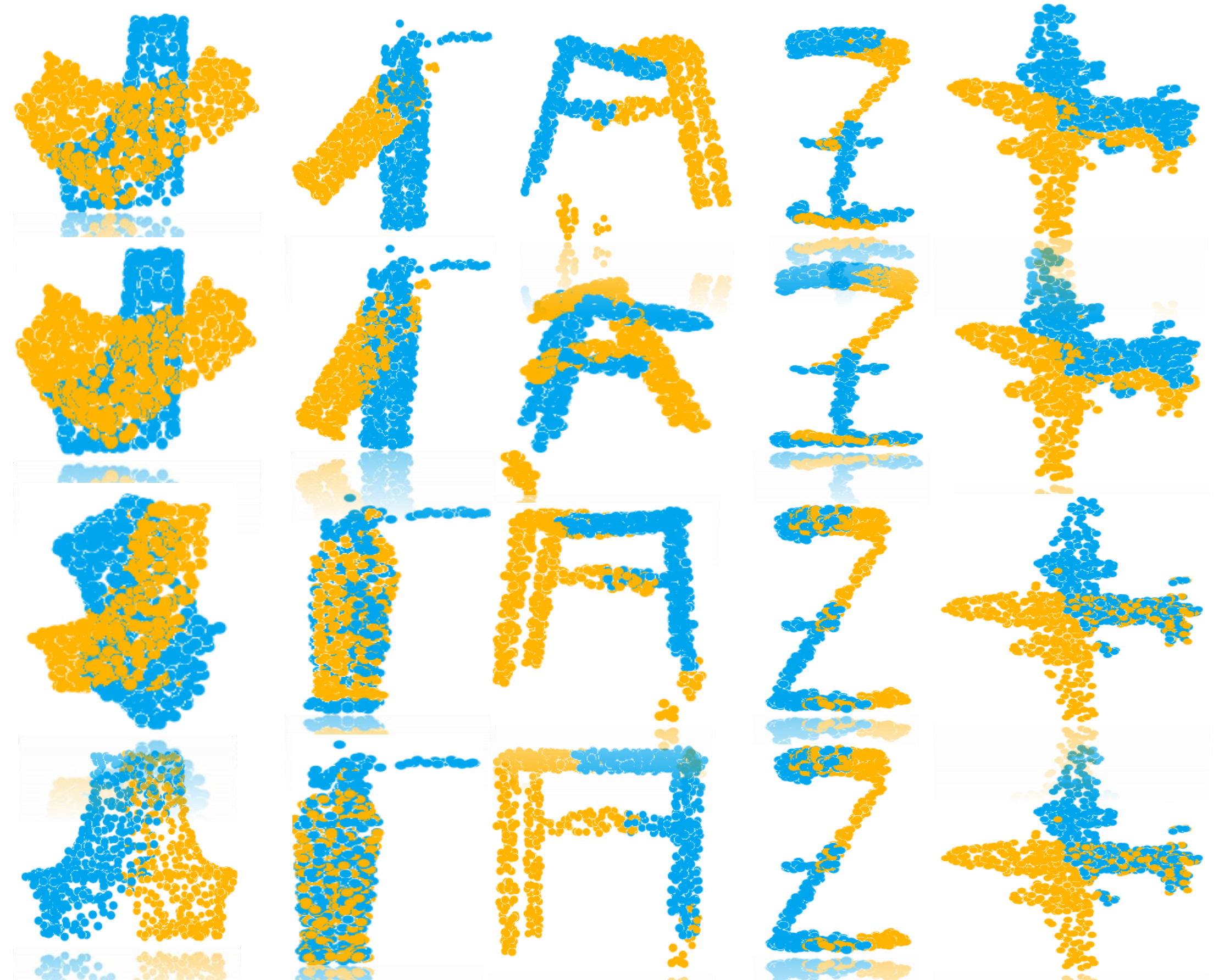}
    \put(-3.5,69.0){\color{black}\footnotesize\rotatebox{90}{\textbf{Input}}}
    \put(-3.5,48.0){\color{black}\footnotesize\rotatebox{90}{\textbf{ICP}}}
    \put(-3.5,25.4){\color{black}\footnotesize\rotatebox{90}{\textbf{RIENet}}}
    \put(-3.5,8.0){\color{black}\footnotesize\rotatebox{90}{\textbf{Ours}}}
    \end{overpic}
	\caption{Registration results of different methods on ModelLoNet with more than 50\% partial overlaps.}
	\label{fig:mnlonet}
 \vspace{-0.3cm}
\end{figure}

\vspace{0.1cm}
\noindent\textbf{Registration Results.} 
Table~\ref{table:mnet} reports registration results on ModelNet40, in which the best results for each criterion are labeled in bold, and the best results by unsupervised methods are underlined.
We compare against the recent unsupervised~\cite{shen2022reliable} and supervised~\cite{wang2019deep,xu2021omnet,huang2021predator,yew2022regtr,yu2021cofinet,qin2022geometric} methods. When compared with unsupervised methods, our \ourmethod outperforms the correspondence-based CEMNet, RIENet and GMM-based UGMM~\cite{huang2022unsupervised} in all metrics under both normal overlap (ModelNet) and low overlap (ModelLoNet) regimes. 
Compared with supervised methods, our approach also achieves competitive results. 
Specifically, our \ourmethod outperforms all previous methods regarding rotation and translation criteria. It is worth noting that RPM-Net~\cite{yew2020rpm} additionally uses surface normals and is trained with transformation information. Despite this, the \ourmethod still performs better. In addition to the quantitative results, Fig.~\ref{fig:mndet} shows results on ModelNet with more than 70.0\% partial overlap. We also offer registration results for ModelLoNet with more than 50.0\% partial overlap in Fig.~\ref{fig:mnlonet}. 
Compared with the recent SOTA unsupervised method RIENet, our \ourmethod recovers the transformation more accurately on the challenging dataset ModelLoNet.

\vspace{0.1cm}
\noindent\textbf{Loss Functions.} We trained our model with different combinations of the local contrastive loss (LC), cross consistency loss (CC), and self-consistency loss (SC), where the experiments were conducted on both ModelNet and ModelLoNet. Table~\ref{tb:loss} shows that the cross-consistency, self-consistency, and local contrastive losses can boost registration precision. Specifically, for a single loss, self-consistency loss archives the best results, and local contrastive loss performs worse on all metrics on both datasets.

\vspace{0.1cm}
\noindent\textbf{Influence of the Number of Clusters.}
We assess the effect of the number of clusters $L$ for ModelNet and ModelLoNet.
We trained \ourmethod with different values of $L$, from 4 to 160, and report the results in Table~\ref{tab:ablation_j}.
\ourmethod achieves the best results with $L{=}64$ on both benchmarks. 
The results are stable for $16 {\leq} L {\leq} 96$. This suggests that the number of clusters has little influence as long as there are ``enough". 

\begin{table}[t]
	\centering
	\caption{Ablation study results of \ourmethod on ModelNet40 with different number of clusters $L$. The best results for each criterion are labeled in bold.}
	\label{tab:ablation_j}
 	\resizebox{0.9\linewidth}{!}{%
	\begin{tabular}{r | c c c | c c c}
	\toprule
	~ &  \multicolumn{3}{c|}{ModelNet} &  \multicolumn{3}{c}{ModelLoNet} \\
	Clusters  & RRE $\downarrow$ & RTE $\downarrow$ & CD $\downarrow$
    & RRE $\downarrow$ & RTE $\downarrow$  & CD $\downarrow$ \\
	\midrule
	4 
    & 1.504 & 0.009 & 0.0366 
    & 4.348 & 0.068 & 0.0452\\ 
    16  
    & 1.439 & 0.007 & 0.0334 
    & 3.713 & 0.057 & 0.0424 \\
    32
    & \bf1.305 & 0.014 & 0.0339 
    & 3.659 & 0.058 & 0.0419\\
    64
    & 1.331 & \bf0.011 & \bf0.0306
    & \bf3.578 & \bf0.069 & \bf0.0416 \\
    96
    & 1.454 & 0.021 & 0.0367
    & 3.598 & 0.070 & 0.0428 \\
    128
    & 1.468 & 0.009 & 0.0310
    & 4.440 & 0.057 & 0.0399\\  
    160
    & 1.530 & 0.009 & 0.0338
    & 4.564 & 0.059 & 0.0422\\
		\bottomrule
	\end{tabular}}
\label{tb:clus}
\end{table}

\vspace{0.1cm}
\noindent\textbf{Importance of Individual Modules.} 
In the registration process, \ourmethod extracts hierarchical correspondences from clusters to points. Therefore, we further explore the efficiency of the hierarchical registration strategy. Table~\ref{tb:cfreg} reports the results on ModelNet and ModelLoNet, where \textit{Cluster}, \textit{Point}, and \textit{Cluster-point} indicate distribution-level, point-level, and distribution-based point-level correspondences, respectively. In the first experiment, we only used distribution-level correspondences for point cloud registration. Unsurprisingly, it performs worse on all metrics, indicating \ourmethod benefits from point-level matching. In the second experiment, we directly predict the point-level correspondences to estimate transformation by performing feature matching. Its performance is still worse than that of the hierarchical registration strategy, further showing the effectiveness of our correspondence prediction strategy.

\begin{table}[t]
	\centering
	\caption{Ablation study of individual modules on ModelNet and ModelLoNet. The best performance is highlighted in bold.}
	\label{tab:ab_clu}
 	\resizebox{0.95\linewidth}{!}{%
	\begin{tabular}{r | c c c | c c c}
	\toprule
	~ &  \multicolumn{3}{c|}{ModelNet} &  \multicolumn{3}{c}{ModelLoNet} \\
	Method  & RRE $\downarrow$ & RTE $\downarrow$ & CD $\downarrow$
    & RRE $\downarrow$ & RTE $\downarrow$  & CD $\downarrow$ \\
	\midrule
	Cluster 
    & 3.932 & 0.033 & 0.0330
    & 6.018 & 0.182 & 0.0463 \\	
    Point  
    & 2.505 & 0.014 & 0.0311
    & 4.264 & 0.096 & 0.0431\\
    Cluster-Point
    & \bf 1.331 & \bf 0.011 & \bf 0.0306
    & \bf 3.578 & \bf 0.069 & \bf 0.0416 \\	    
	\bottomrule
	\end{tabular}}
\label{tb:cfreg}
\vspace{-0.3cm}
\end{table}


\section{Conclusion}
This paper presents a distribution consistency-based unsupervised deep probabilistic registration framework. One of the advantages of this method is that it extends the probabilistic registration to handle point cloud registration with partial overlaps by adopting the Sinkhorn algorithm to predict distribution-level correspondences. Moreover, we propose self-consistent, cross-consistent, and local-contrastive losses to train feature extractors in an unsupervised manner. Experiments demonstrate that the proposed algorithm achieves the best performance.



{\small
\bibliographystyle{ieee_fullname}
\bibliography{egbib}
}
\newpage
\appendix
\twocolumn[
\begin{@twocolumnfalse}
\section*{\centering{Unsupervised Deep Probabilistic Approach for Partial Point Cloud Registration \\ --Supplementary Material--}}
\vspace{1cm}
\end{@twocolumnfalse}]

\section{Appendix}
In this supplementary material, we first describe the detailed feature extractors in Sec.~\ref{sec:afe}, then we provide the details of the Transformer in Sec.~\ref{sec:attn}, followed by solving Eq.~\eqref{eq:sgamma} in Sec.~\ref{sec:opt}. We also give the definitions of evaluation metrics in Sec.~\ref{sec:metric}.
Finally, we provide more registration results in Sec.~\ref{sec:m_res}.

\subsection{Feature Extractor}~\label{sec:afe}
Our \ourmethod adopts a KPConv~\cite{thomas2019kpconv}-based encoder-decoder architecture for feature
extraction, where we add a lightweight Transformer for context aggregation.
The configurations of KPConv and ResBlock are the same as in~\cite{huang2021predator}.
On 3DMatch, following~\cite{qin2022geometric}, we first downsample the input point clouds with
a voxel size of 2.5cm, then send the downsampled point clouds into the feature extractor.
The detailed network configurations are shown in Table~\ref{tb:arch}.

\begin{table}[!hbt]
    \centering
    \caption{Network architecture for 3DMatch and ModelNet.}
    \label{tb:arch}
    \resizebox{1\linewidth}{!}{%
    \begin{tabular}{c|c c}
    \toprule
    Stage & 3DMatch & ModelNet\\
    \midrule
    \multirow{2}{*}{1}    
    & KPConv($1{\rightarrow} 64$) & KPConv($1{\rightarrow} 256$) \\
    & ResBlock($64{\rightarrow} 128$) & ResBlock($256{\rightarrow} 256$) \\
    \midrule
    \multirow{3}{*}{2}
    & ResBlock($64{\rightarrow} 128$, strided) & ResBlock($256{\rightarrow} 512$, strided) \\
    & ResBlock($128{\rightarrow} 256$) & - \\
    & ResBlock($256{\rightarrow} 256$) & - \\
    \midrule
    \multirow{3}{*}{3}
    & ResBlock($256{\rightarrow} 256$, strided) & ResBlock($512{\rightarrow} 512$, strided) \\
    & ResBlock($256{\rightarrow} 512$) & ResBlock($512{\rightarrow} 512$) \\
    & ResBlock($512{\rightarrow} 512$) & - \\
    \midrule
    \multirow{3}{*}{4}
    & ResBlock($512{\rightarrow} 512$, strided) & ResBlock($1024{\rightarrow} 1024$, strided) \\
    & ResBlock($512{\rightarrow} 1024$) & - \\
    & ResBlock($1024{\rightarrow} 1024$) & - \\
    \midrule
    \multirow{2}{*}{5} 
    & -                                       & ResBlock($1024{\rightarrow} 1024$, strided) \\
    & -                                       & ResBlock($1024{\rightarrow} 1024$) \\
    \midrule
    6 & -                                   & ResBlock($1024{\rightarrow} 1024$, strided) \\
    \midrule
    \multirow{2}{*}{7} 
    & Conv1D($1024{\rightarrow}256$) & Conv1D($1024{\rightarrow}256$) \\
    & Transformer($256{\rightarrow}256$) & Transformer($256{\rightarrow}256$) \\
    \midrule
    \multirow{2}{*}{8}
    & NearestUpsampling & NearestUpsampling \\
    & UnaryConv($1537{\rightarrow}512$) 
    & UnaryConv($1537{\rightarrow}512$) \\
    \midrule
    \multirow{2}{*}{9}
    & NearestUpsampling & NearestUpsampling \\
    & UnaryConv($768{\rightarrow}256$) 
    & UnaryConv($768{\rightarrow}512$) \\
    \midrule
    10  & Linear($512{\rightarrow}257$) & Linear($256{\rightarrow}129$) \\
    \bottomrule
    \end{tabular}
    }
\end{table}
\subsection{Transformer}~\label{sec:attn}
The transformer is composed of three main components: self-attention, positional encoding, and cross-attention. Geometric self-attention is utilized to capture long-range dependencies, while positional encoding assigns intrinsic geometric properties to each point feature, thereby increasing differentiation among features in areas where they may be indistinct. The cross-attention module leverages the connections between the source and target point clouds, enabling the encoding of contextual information between partially overlapping point clouds. The individual parts will be described in detail below.

\paragraph{Self-Attention.} 
We use the geometric self-attention provided in GeoTransformer~\cite{qin2022geometric} for self-attention.

\paragraph{Positional Encoding.} 
Following~\cite{mei2022overlap}, incorporating a positional encoding approach, which imparts intrinsic geometric attributes to individual point features through the inclusion of unique positional information, improves differentiation among point features in less distinctive regions. To begin with, we choose the $k=10$ nearest neighbors $\mathcal{K}_i$ of $\bar{\bm{p}}_i^s$ and calculate the centroid $\bar{\bm{p}}^s_c=\sum_{i=1}^{\bar{N}_s}\bar{\bm{p}}^s_i$ of $\bar{\bm{\mathcal{P}}}^s$, where $\bar{\bm{p}}^s_i$ and $\bar{\bm{p}}^s_j$ represent two superpoints of $\bar{\bm{\mathcal{P}}}^s$.
For each $\bar{\bm{p}}^s_x\in \mathcal{K}_i$, we denote the angle between two vectors $\bar{\bm{p}}^s_i-\bar{\bm{p}}^s_c $ and $ \bar{\bm{p}}^s_x-\bar{\bm{p}}^s_c$ as $\alpha_{ix}$.
The position encoding $\bar{\bm{g}}^s_i$ of $\bar{\bm{p}}_i$ is defined as follows: 
\begin{equation}\label{eq:pos}
	\bar{\bm{g}}^s_i= \varphi\left(\|\bar{\bm{p}}^s_i-\bar{\bm{p}}^s_c\|_2\right) +\max_{x\in \mathcal{K}_i}\{\phi\left(\alpha_{ix}\right)\},
\end{equation}
where $\varphi$ and $\phi$ are two MLPs, and each MLP consists of a linear layer and one ReLU nonlinearity function.

\paragraph{Cross-Attention.} 
Let ${^{(l)}\bar{\bm{\mathcal{F}}}^s}$ be the intermediate representation for $\bar{\bm{\mathcal{P}}}$ at layer $l$ and let ${^{(0)}\bar{\bm{\mathcal{F}}}^s}{=}\{\bar{\bm{g}}^s_i{+}\bar{\bm{f}}^s_i\}_{i=1}^{\bar{N}_s}$. We use a multi-attention layer consisting of four attention heads to update the ${^{(l)}\bar{\bm{\mathcal{F}}}^s}$ via
\begin{equation}\label{eq:sim}
	\begin{aligned}
			&\bm{S}^s{=} {^{\left(l\right)}\bm{W}}_1{^{(l)}\bar{\bm{\mathcal{F}}}^s} {+} {^{(l)}\bm{b}}_1, \bm{K}^{t} {=} {^{\left(l\right)}\bm{W}}_2 {^{(l)}\bar{\bm{\mathcal{F}}}^t} {+} {^{(l)}\bm{b}}_2, \\
			&\bm{V}^{x}{=} {^{\left(l\right)}\bm{W}}_3{^{(l)}\bar{\bm{\mathcal{F}}}^t} {+} {^{(l)}\bm{b}}_3, \bm{A} {=} \mbox{softmax}\left(\frac{{\bm{S}^s}^\top \bm{K}^{t}}{\sqrt{b}}\right), \\
			&{^{(l+1)}\bar{\bm{\mathcal{F}}}^s} {=}{^{(l)}\bar{\bm{\mathcal{F}}}^s} {+} {^{(l)}h}\left(\bm{A}\bm{V}^x\right).
		\end{aligned}
\end{equation}
Here, $^{(l)}h\left(\cdot\right)$ is a three-layer fully connected network consisting of a linear layer, instance normalization, and a LeakyReLU activation. The same attention module is also simultaneously performed for all points in point cloud $\bar{\bm{\mathcal{P}}}^t$. 
The final outputs of attention module are $\bar{\bm{\mathcal{F}}}^s$ for $\bar{\bm{\mathcal{P}}}^s$ and $\bar{\bm{\mathcal{F}}}^t$ for $\bar{\bm{\mathcal{P}}}^t$. The latent features $\bar{\bm{\mathcal{F}}}^s$ have the knowledge of $\bar{\bm{\mathcal{F}}}^t$ and vice versa.

\paragraph{Overlap Score.} 
After computing $\bar{\bm{\mathcal{F}}}^s$ and $\bar{\bm{\mathcal{F}}}^t$, a network acts on them to extract overlap scores $\bar{\bm{O}}^s=\{\bar{\bm{o}}^s_i\in [0, 1]\}_{i=1}^{\bar{N}}$ and $\bar{\bm{O}}^t=\{\bar{\bm{o}}^t_j\in [0, 1]\}_{j=1}^{\bar{M}} \in [0, 1]$ for $\bar{\bm{\mathcal{P}}}^s$ and $\bar{\bm{\mathcal{P}}}^t$, respectively, to identify the overlapping regions~\cite{huang2021predator}.
The overlap scores and features are sent to the decoder, which outputs the point-wise feature descriptor $\bm{\mathcal{F}}^{s}{\in}\mathbb{R}^{N_s\times d}$ and $\bm{\mathcal{F}}^t{\in}\mathbb{R}^{N_t\times d}$ and overlap scores $\bm{O}^s{=}\{o_i^s\}{\in}\mathbb{R}_+^{N_s}$ and $\bm{O}^t=\{o_j^t\}{\in}\mathbb{R}_+^{N_t}$. 
$d$ is the dimension of features.

\subsection{Optimization}\label{sec:opt}
Now, we introduce how to address the optimization objective presented in Eq.~\eqref{eq:sgamma} of the main paper:
\vspace{-.2cm}
\begin{equation}\label{eq:agamma}
\begin{aligned}
& \min_{\bm{\gamma}}\sum_{i,j}\bm{\gamma}_{ij}\|\bm{p}_i-\bm{\mu}_j\|^2_2,\\
& \mbox{s.t.,}~ \sum_i\bm{\gamma}_{ij} {=} N\bm{\pi}_j, \sum_j\bm{\gamma}_{ij} {=} 1, \bm{\gamma}_{ij}\in[0, 1].
\end{aligned}
\end{equation}
The constraint $\sum_j\bm{\gamma}_{ij}{=}1$ is imposed based on the property of probability that the sum of all probabilities for all possible events is equal to one. The constraint $\sum_i\bm{\gamma}_{ij} {=} N\bm{\pi}_j$ represents the mixture weights' constraints.

Let $\bm{\Gamma}=\frac{\bm{\gamma}}{N}$ with elements defined as $\Gamma_{ij}=\frac{\gamma_{ij}}{N}$. By replacing the variable $\bm{\gamma}$ with $\bm{\Gamma}$ in Eq. (\ref{eq:agamma}), the joint objective can be formulated as an optimal transport (OT) problem~\cite{peyre2019computational} as
\begin{equation}\label{eq:poster}
    \min_{\bm{\Gamma}} \left<\bm{\Gamma}, \bm{D}\right>, ~ \mbox{s.t.} ~ \bm{\Gamma}^\top\bm{1}_N=\bm{\pi}, \bm{\Gamma}\bm{1}_L=\frac{1}{N}\bm{1}_N.
\end{equation}
While the minimization of Eq. (\ref{eq:poster}) can be solved in polynomial time as a linear program, it becomes challenging when dealing with millions of data points and thousands of classes as traditional algorithms do not scale well \cite{cuturi2013sinkhorn}. To overcome this limitation, we utilize an efficient version of the Sinkhorn-Knopp algorithm\cite{cuturi2013sinkhorn}.
This requires the following regularization term:
\begin{equation}\label{eq:opt}
	\begin{aligned}
		& \min_{\bm{\Gamma}} \left<\bm{\Gamma}, \bm{D}\right> - \epsilon H\left(\bm{\Gamma}\right), \\
		& \mbox{s.t.} ~ \bm{\Gamma}^\top\bm{1}_N=\bm{\pi}, \,\,\,
		\bm{\Gamma}\bm{1}_L=\frac{1}{N}\bm{1}_N,
	\end{aligned}
\end{equation}
where $H\left(\bm{\Gamma}\right)=\left<\bm{\Gamma},\log\bm{\Gamma}-1\right>$ represents the entropy of $\bm{\Gamma}$, and $\epsilon > 0$ is a regularization parameter. When $\epsilon$ is very large, optimizing Eq.~\eqref{eq:opt} is equivalent to optimizing Eq.~\eqref{eq:poster}, but even for moderate values of $\epsilon$, the objective function tends to have approximately the same optimal solution \cite{cuturi2013sinkhorn}. Choosing the appropriate value of $\epsilon$ involves a trade-off between convergence speed and proximity to the original transport problem \cite{cuturi2013sinkhorn}. In our scenario, a fixed value of $\epsilon$ is suitable since our focus is on obtaining the final clustering and representation learning outcomes rather than solving the transport problem exactly. The solution to Eq.~(\ref{eq:opt}) can be expressed as a normalized exponential matrix, as stated in \cite{cuturi2013sinkhorn},
\begin{equation}\label{eq:gamma}
	\bm{\Gamma} = \mbox{diag}\left(\bm{\mu}\right)\exp\left(\bm{D}\big/\epsilon\right)\mbox{diag}\left(\bm{\nu}\right),
\end{equation}
where $\bm{\mu}=(\mu_1, \mu_2,\cdots,\mu_N)$ and $\bm{\nu}=(\nu_1,\nu_2,\cdots,\nu_L)$ are renormalization vectors in $\mathbb{R}^N$ and $\mathbb{R}^L$. Iterating the updates via $\bm{\mu}_i=\left[\exp\left(\bm{D}\big/\epsilon\right)\bm{\nu}\right]^{-1}_i$ and $\bm{\nu}_j=\left[\exp\left(\bm{D}\big/\epsilon\right)^\top\bm{\mu}\right]^{-1}_j$ with initial values $\bm{\mu}=\frac{1}{N}\bm{1}_N$ and $\bm{\nu}=\bm{\pi}$, respectively, yields the vectors $\bm{\mu}$ and $\bm{\nu}$. Although any distribution can be used for the initialization of $\bm{\mu}$ and $\bm{\nu}$, setting them as the constraints results in faster convergence~\cite{cuturi2013sinkhorn}. In our experiments, we used 20 iterations as it worked well in practice. After solving Eq. (\ref{eq:gamma}), we obtain the probability matrix $\bm{\gamma}$ as
\begin{equation}\label{eq:soft_labels}
    \bm{\gamma}=N\cdot\bm{\Gamma}.
\end{equation}
Eqs.~\eqref{eq:mgamma}, ~\eqref{eq:feature}, and \eqref{eq:point} can be solved in a similar way.

\subsection{Metrics}\label{sec:metric}
Following Predator~\cite{huang2021predator} and CoFiNet~\cite{yu2021cofinet}, we use three metrics, \textit{Registration Recall} ($RR$), \textit{Relative Rotation Error} ($RRE$), and \textit{Relative Translation Error} ($RTE$), to evaluate the performance of the proposed registration algorithm. RRE and RTE are respectively defined as 
\begin{equation}
\begin{aligned}
RRE &= \arccos\left(\frac{\textbf{Tr}\left(\bm{R}^\top\bm{R}^\star\right)-1}{2}\right),\\
RTE &=\|\bm{t}-\bm{t}^\star\|_2,
\end{aligned}
\end{equation}
where $\bm{R}^\star$ and $\bm{t}^\star$ denote the ground-truth rotation matrix and the translation vector, respectively. \textit{Registration Recall} (RR), the fraction of point cloud pairs whose root mean square error (RMSE) of transformation is smaller than a certain threshold (i.e., $RMSE<0.2m$).
Specifically, we denote the set of ground truth correspondences as $\mathcal{H}$ and the estimated transformation $T$, their root mean square error are calculated as:
\begin{equation}
    \mbox{RMSE} = \sqrt{\frac{1}{|\mathcal{H}|}\sum_{(\bm{p},\bm{q})\in\mathcal{H}}\|T(\bm{p})-\bm{q}\|_2^2}.
\end{equation}
Follow~\cite{huang2021predator}, Chamfer distance (CD) is used to measure the registration quality on ModelNet40. We use the modified Chamfer distance metric:
\begin{equation}
\begin{aligned}
    CD\left(\mathcal{\bm{P}},\mathcal{\bm{Q}}\right)&{=}
    \frac{1}{|\mathcal{\bm{P}}|}\sum_{\bm{p}\in\mathcal{\bm{P}}}\min_{\bm{q}\in\mathcal{\bm{Q}}}\|T\left(\bm{p}\right){-}\bm{q}\| \\
    &
    {+}\frac{1}{|\mathcal{\bm{Q}}|}\sum_{\bm{q}\in\mathcal{\bm{Q}}}\min_{\bm{p}\in\mathcal{\bm{P}}}\|T\left(\bm{p}\right){-}\bm{q}\|, 
\end{aligned}
\end{equation}
where $\mathcal{\bm{P}}$ and $\mathcal{\bm{Q}}$ are
input source and target point clouds.

\subsection{More Results}\label{sec:m_res}
\begin{figure}[t]
\centering 
\begin{overpic}[width=1.0\columnwidth]{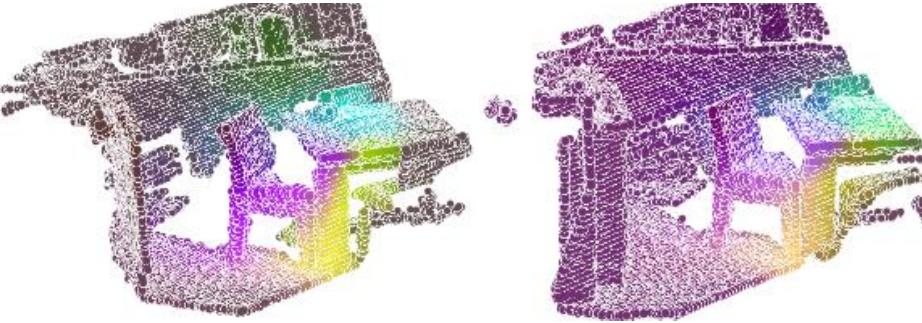}
 \put(20,36.0){\color{black}\footnotesize{\textbf{Source}}}
 \put(70,36.0){\color{black}\footnotesize{\textbf{Target}}}
\end{overpic}
\caption{Coloring the points using learned GMM labels.} \label{fig:color}
\end{figure}

\paragraph{Visualize the Gaussian mixtures.}
Fig.~\ref{fig:color} shows the visual results by coloring the points using GMM labels. We use different colors to differentiate clusters.

\paragraph{KITTI results.}
Table~\ref{tab:kitti} shows the generalization results from 3DMatch to KITTI.
\ourmethod outperforms baselines, showing its robustness and generalization.
\begin{table}[!hbt]
\centering
\caption{Results of generalization from 3DMatch to KITTI.}\label{tab:kitti}
\resizebox{0.99\linewidth}{!}
{%
\begin{tabular}{l c c c c}
\toprule
Method & RTE($\uparrow$) & RRE($\uparrow$) & Success($\uparrow$) & Time ($\downarrow$) \\
\toprule
Predator     & 16.5 & 1.38 & 46.13 & 0.44 \\
SGP          & 13.8 & 0.49 & 62.22 & \bf 0.12 \\
UDPReg(Ours) & \bf8.81 & \bf0.41 & \bf64.59 & 0.26 \\
\bottomrule
\end{tabular}
 }
\end{table}

\paragraph{Complexity Analysis.} 
{\footnotesize$O(N{\times} L),L{<}N$} complexity for clustering and $O(N^2)$ for attention represents the memory bottleneck of UDPReg. $N, L$ are point and cluster numbers, respectively. 
Table~\ref{tab:kitti} reports the inference time on a Tesla V100 GPU (32G) and two Intel(R) 6226 CPUs.

\end{document}